# A Self-Organizing Swarm Control Model for Underwater Robots Based on Turing-Type Reaction-Diffusion Principles

Takeshi Ishida[1*]

[1]Department of Ocean Mechanical Engineering, National Fisheries University;
Shimonoseki, 759-6595, Japan.

* Takeshi Ishida
Address correspondence to: ishida@fish-u.ac.jp

**Abstract**

In underwater environments, Global Positioning System (GPS) signals and high-speed radio communication are unavailable, and severe constraints exist on communication bandwidth and visibility; consequently, applying conventional swarm formation methods, such as the Boids model, which require the relative positions and velocity vectors of neighboring vehicles, is difficult. This study proposes a decentralized swarm control model that operates without a leader or absolute coordinates, relying solely on the number of received signals from two communication channels with different transmission ranges (inner and outer neighborhoods). The proposed model extends a cellular automaton model based on Turing pattern formation principles to continuous space and is grounded in the fundamental concept that swarms form as robots move toward regions of higher potential calculated from the difference in received signal counts. Specifically, stable behavior is determined through temporal averaging of potential values, hysteresis-based state evaluation, stepwise turning based on temporal differences, and vector composition with an external directional command. Two-dimensional simulation results demonstrated that autonomous swarm formation and control of swarm spatial dispersion via a morphological parameter are achievable, confirming robustness in maintaining swarms and high polarization even under communication dropouts where the communication success rate decreases to approximately 0.7.

## 1. Introduction

### 1.1 Current Status and Challenges of Underwater Robots

In recent years, the importance of underwater robots has been increasing in various fields, including marine environmental surveys, seabed resource exploration, the inspection and maintenance of underwater structures, port facility management, and fishery and aquaculture monitoring. Because the underwater domain presents a harsh environment in which humans cannot safely operate for extended periods, there is a strong demand for robotic technologies capable of complementing or replacing human diving operations to survey wide marine areas continuously and efficiently.

Representative examples of underwater robots include Remotely Operated Vehicles (ROVs) and Autonomous Underwater Vehicles (AUVs). Furthermore, compact and relatively low-cost ROVs—commonly referred to as "underwater drones"—have recently become commercially available. Compared with conventional large-scale ROVs, these vehicles are easier to deploy and have the potential to reduce reliance on human underwater labor. However, most current underwater drones fundamentally require remote control by a human operator. Because continuous attention is required even to maneuver a single vehicle safely and appropriately, it is difficult for a single operator to control multiple vehicles simultaneously.

Meanwhile, recent developments include hybrid vehicles combining AUV and ROV functions, resident underwater vehicles capable of recharging and transferring data at seabed docking stations, and cooperative systems for multiple AUVs [1]. Although increasing autonomy and multi-vehicle coordination have been partially achieved, the individual control of single robots remains the predominant approach in practical operations. Under severe underwater communication constraints, a versatile control method for flexibly and stably coordinating numerous robots has not yet been fully established.

In terrestrial and aerial environments, Global Positioning System (GPS) data and high-speed radio communication are readily available. Consequently, individual robots can share their positions, velocities, and headings to act cooperatively under commands from a central controller. In contrast, GPS signals do not penetrate water, and the effective range of conventional radio communication is severely limited underwater. Although acoustic communication can cover relatively long distances, it suffers from narrow bandwidth and is susceptible to transmission delays, packet loss, multipath propagation, and noise [2]– [7]. Moreover, depending on water depth, turbidity, and lighting conditions, continuous visual recognition of neighboring robots using cameras is also difficult.

Therefore, cooperative control of underwater robot swarms requires a control method that can autonomously form and maintain a swarm using only limited information, without assuming continuous access to the detailed positions, velocity vectors, or attitudes of other robots. Furthermore, a distributed algorithm is required that avoids excessive reliance on a central controller or a specific leader and maintains overall swarm functionality even under partial communication loss.

### 1.2 Related Work on Swarm Control of Underwater Robots

#### 1.2.1 Research Trends in Underwater Multi-Robot Systems

Research on coordinating multiple autonomous underwater robots has developed to improve operational efficiency through multi-robot cooperation. Such cooperation enables tasks that are difficult for a single ROV or AUV, including wide-area exploration, seabed mapping, marine environmental monitoring, and structural inspection. Early studies primarily focused on formation control, in which multiple AUVs move while maintaining a predefined geometric configuration. Subsequently, the research scope has expanded to broader underwater multi-robot systems, addressing communication constraints, obstacle avoidance, fault tolerance, task allocation, and cooperative exploration.

Das et al. [2] reviewed early studies on multi-AUV cooperative exploration and formation control, identifying formation keeping, path following, communication constraints, fault detection, and collision avoidance as key research challenges. Connor et al. [3] surveyed underwater swarm robotics in terms of robot design, perception, localization, communication methods, centralized and decentralized control, and collective behavior. Yang et al. [4] categorized AUV formations from three perspectives: vehicle kinematic performance, formation control schemes, and communication capabilities. Zhou et al. [5] classified underwater multi-robot systems based on task space, motion space, and measurement space. Furthermore, Yan et al. [6] compared various multi-AUV formation control methods. Cai et al. [7] also discussed the potential of cooperative artificial intelligence for underwater robot swarms across cooperative communication, navigation, information fusion, decision-making, and task allocation.

A common consensus among these survey papers is that communication and localization constraints are significantly more severe for underwater robot swarms than for terrestrial or aerial ones [2]– [7]. Consequently, it is difficult for individual robots to continuously share accurate information regarding the positions, velocities, and attitudes of other vehicles. For the practical deployment of underwater robot swarms, developing a decentralized control scheme capable of operating under imperfect communication conditions remains a critical challenge.

### 1.2.2 Formation and Cooperative Control of Multiple AUVs

A representative method for the cooperative control of multiple AUVs is the leader–follower approach. Cui et al. [8] proposed a formation control scheme in which follower vehicles track a leader, specifically targeting underactuated AUVs (i.e., vehicles having fewer independent actuators—such as thrusters, rudders, and fins—than degrees of freedom). The leader–follower approach is relatively simple and allows the target formation geometry to be clearly defined. However, leader failure or communication loss can compromise the entire group, distinguishing this approach from fully decentralized control schemes.

Hou and Cheah [9] demonstrated that multi-AUV formation control can be achieved using relatively simple control laws, such as proportional–derivative (PD) and adaptive PD control. Zhen et al. [10] combined the virtual structure method with artificial potential fields to address 3D position and attitude control, as well as obstacle avoidance, for AUV formations. Although these methods can maintain multiple robots in stable formations, they generally require information on relative positions, distances, bearings, and velocities with respect to other vehicles, or predefined formation geometries.

In recent years, studies explicitly addressing the imperfections of underwater communication have also progressed. Li et al. [11] investigated multi-AUV formation control under conditions involving communication delays, packet discreteness, and packet loss. Furthermore, Fang et al. [12] proposed a method using multi-agent adversarial imitation learning to train multiple AUVs in formation control and obstacle avoidance. Dunbabin et al. [13] conducted cooperative control experiments combining underwater robots with sensor networks, demonstrating that multi-robot collaboration can expand the capabilities of a single vehicle.

These studies share a commonality with the present work in coordinating multiple robots underwater. However, most existing studies rely on several assumptions, such as maintaining predefined formations, tracking a leader, estimating the relative positions of other vehicles, predicting states under communication delays, or utilizing pre-trained policies. In contrast, this study proposes a control scheme capable of forming and maintaining a swarm without designating a specific leader or requiring detailed position coordinates and velocity vectors of other vehicles.

### 1.2.3 Swarm Control Based on Self-Organization

Decentralized control based on self-organization is an important research approach to robot swarm formation. Reynolds [14] demonstrated that collective motion resembling bird flocks or fish schools can be generated using three local rules: separation, alignment, and cohesion. Commonly known as the Boids model, this framework forms the foundation of swarm control research. Furthermore, Olfati-Saber [15] theoretically formulated flocking in multi-agent systems using graph theory, potential functions, and velocity consensus.

In these self-organizing models, ordered collective motion emerges without a central controller through repeated local interactions among neighboring agents. However, conventional Boids models require each agent to acquire the relative positions, distances, headings, and velocities of surrounding agents. In underwater environments with poor visibility and limited communication bandwidth, continuously acquiring such information is often difficult.

Additionally, in underwater environments, Berlinger et al. [16] investigated local interactions using a swarm of miniature fish-inspired robots. Without relying on centralized control or shared global coordinates, their system achieved 3D collective behaviors such as aggregation, dispersion, milling, and search through local perception and implicit coordination. This study demonstrates that complex collective behaviors of underwater robots can emerge from simple local interactions.

Furthermore, Sahu and Subudhi [17] proposed a flocking control method that maintains connectivity among multiple AUVs using fuzzy potential functions. Zhen et al. [10] similarly applied artificial potential fields to AUV formation control. These studies relate to the present work in that they utilize potential functions to stably maintain a swarm. However, conventional potential field methods compute potential values or control inputs based on inter-robot relative distances, bearings, or position coordinates.

### 1.2.4 Self-Organizing Models Based on Turing Patterns

Turing [18] demonstrated that spatial patterns, such as spots and stripes, can spontaneously emerge from a uniform state through the interaction between local activation and long-range inhibition. This concept forms the theoretical foundation for reaction-diffusion models describing biological morphogenesis. Ishida [19] formulated the fundamental concept of Turing patterns as a cellular automaton by summing cell states within a short-range inner neighborhood and a wider outer neighborhood. This approach was shown to generate and control diverse self-organized patterns in addition to spots, stripes, and moving patterns. Furthermore, Ishida [20] applied this self-organizing model to the control of modular swarm robots. Without using a coordinate system or a designated leader module, the robot group achieved aggregation, locomotion, and shape maintenance solely through local information exchange with adjacent modules. The present study applies this concept to the swarm control of underwater robots moving freely in continuous space.

## 1.3 Research Objectives

This study aims to establish the foundation of a decentralized swarm control model designed to function under limited underwater communication conditions, where each robot determines its behavior based on local received information and internal states. The proposed model neither assigns a specific leader nor assumes the use of an absolute coordinate system; additionally, individual robots do not need to acquire the relative positions, distances, headings, or velocity vectors of other vehicles. For swarm formation, the only information each robot acquires from others is the number of received signals across two communication channels with different transmission ranges. Behavioral decisions are made using these signal counts, along with the robot's own heading, past potential values, internal states, and a common directional command received from an external source.

The fundamental concept of this study is to extend the self-organizing models proposed by Ishida [19] and Ishida [20] to an underwater robot swarm moving freely in continuous space. Specifically, Ishida [19] formulated Turing pattern formation principles using cellular automata, and Ishida [20] subsequently applied this model to modular swarm robots.

Each robot is assumed to have two communication ranges: a relatively narrow inner neighborhood and a wider outer neighborhood. Each robot counts the signals received within these ranges and calculates a potential value at its current position based on the weighted difference between the inner and outer signal counts. Using this potential value as an indicator, a robot can evaluate its relative status within the swarm without acquiring absolute coordinates or detailed positional information of other vehicles.

Furthermore, this study does not assume that each robot directly measures the spatial gradient of the potential value. Each robot stores the potential value from the previous time step and determines whether this value increases or decreases as it moves. If the potential value increases, the robot generally maintains its current heading; if it decreases, the robot probabilistically changes direction. This method is analogous to the principle of chemotaxis [21], in which bacteria alter their frequency of directional changes in response to temporal variations in chemical concentration.

Thus, the proposed model aims to self-organize collective motion with a specific order across the entire swarm. This is achieved through individual robots continuously making local decisions based solely on limited information. Furthermore, this study investigated a control method capable of moving the entire swarm in a designated direction while maintaining its shape. This method combines an external directional command with the intrinsic behavior of maintaining the swarm.

To verify this fundamental principle, a simulation model was developed to simulate multiple underwater robots moving in two-dimensional space. This model evaluates the following three aspects: 1) whether randomly deployed robots can autonomously form a swarm, 2) whether varying parameters can control swarm morphologies, such as the number of swarms, radius, and participation rate, and 3) whether the swarm can maintain its shape and heading even when communication signals or commands are partially lost. These evaluations aim to determine the robustness of the proposed model against the inevitable communication uncertainties of actual underwater environments.

This study does not aim to immediately finalize a physical control system for underwater robot swarms. Instead, the primary objective is to present a simple and interpretable swarm control principle that functions under severely limited information conditions and to verify its fundamental effectiveness through simulations. Future work must address expansions into three-dimensional space, the introduction of ocean currents and obstacles, and the addition of collision avoidance functions. Furthermore, practically viable communication methods must be selected, and task control for distributing multiple operations must be introduced.

## 2. Underwater Robot Swarm Control Model

### 2.1 Overview and Characteristics of the Swarm Control Model

This model demonstrates that robot swarm formation is possible using only two limited types of underwater communication (one reaching relatively far and the other reaching only nearby) by applying the algorithm of the Ishida model [19], which creates Turing patterns using neighborhood range counts.

It is assumed that each robot possesses an ID and emits two types of signals. One signal reaches relatively distant robots (corresponding to the outer neighborhood in the Ishida model), and the other reaches only nearby robots (corresponding to the inner neighborhood in the Ishida model). Swarm formation is achieved using a potential value calculated from the decision formula of the Ishida model [19] based on the number of signals received by each robot. Figure 1 shows an overview of the model.

This model self-organizes an autonomous swarm using only the number of received signals (a scalar quantity), without requiring the detailed position coordinates and velocity vectors of other vehicles in an underwater environment with poor visibility and severely limited communication bandwidth. The advantage of this algorithm lies in eliminating the calculation process of relative coordinates (distance and bearing) of other vehicles, which is essential in conventional swarm control methods such as the Boids model. Furthermore, to suppress hypersensitive responses to temporary fluctuations in signal reception caused by communication dropouts or missed commands, this algorithm introduces the temporal averaging of received signal counts, hysteresis in state evaluation, and stepwise directional changes in response to potential decreases. The following sections detail the algorithm for the swarm control model of the autonomous underwater robots (underwater drones) proposed in this study.

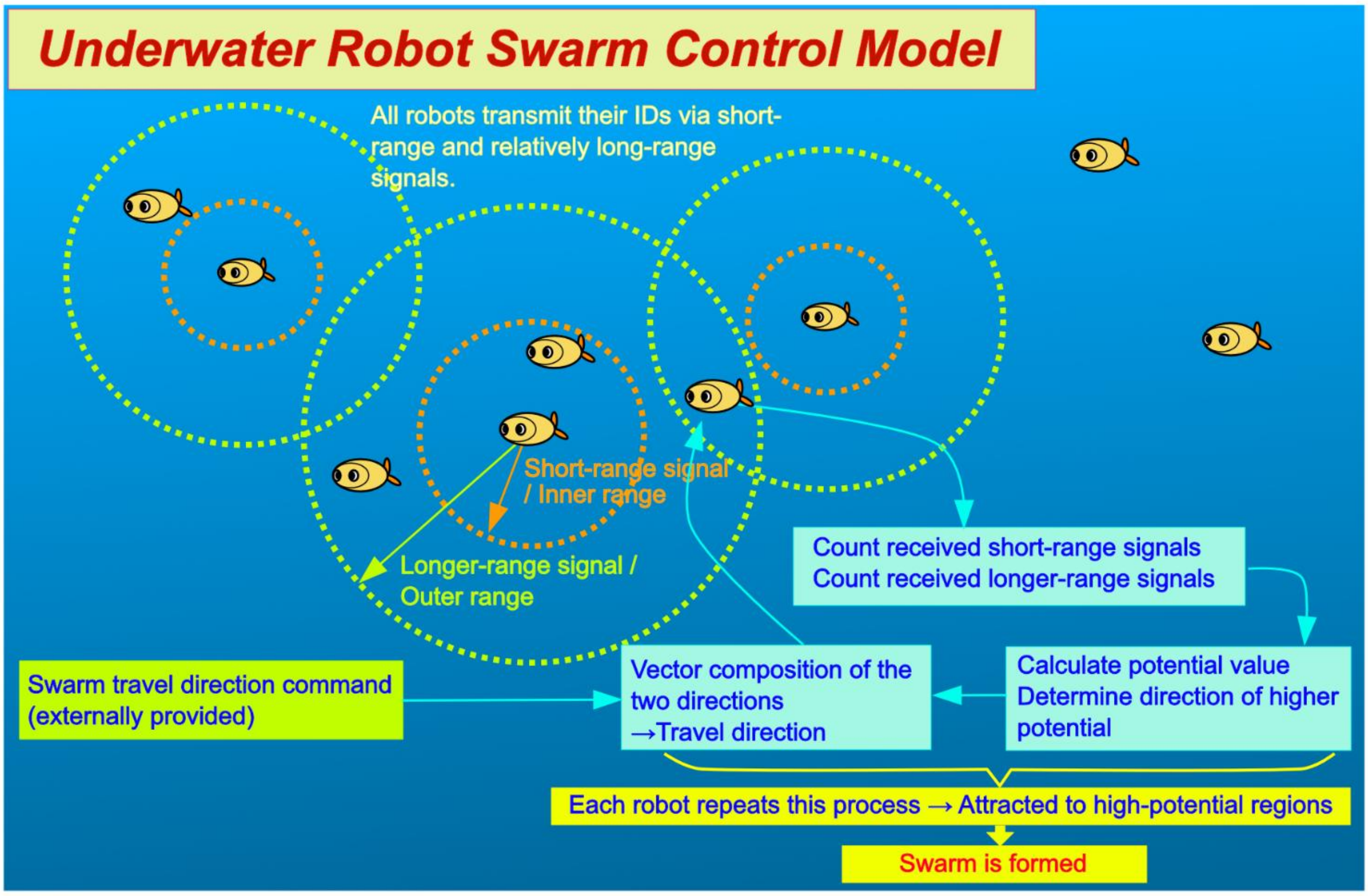

**Figure 1. Overview of the underwater robot swarm control model.** It is assumed that each robot possesses an ID and emits two types of signals. One signal reaches relatively distant robots, while the other reaches only nearby robots. Swarm formation is achieved using a potential value calculated from the decision formula of the Ishida model [19] based on the number of signals received by each robot.

### 2.2 Communication Model for Swarm Control

This model extends the model proposed by Ishida [19], which applies the emergence mechanism of Turing patterns, to the spatial motion control of underwater drones. Each robot is assumed to be capable of receiving two types of signals with different transmission ranges. In this model, these ranges are idealized as circular neighborhoods with radii $R_{inner}$ (inner neighborhood) and $R_{outer}$ (outer neighborhood), where $R_{inner} < R_{outer}$. The outer neighborhood encompasses the inner neighborhood, and the host robot itself is excluded from the received signal count. The propagation characteristics of specific communication devices are not modeled, leaving the selection of physical implementations such as optical or acoustic communication for future work. For practical applications, blue-green LEDs, underwater optical wireless communication (UOWC), or short-range radio frequencies are conceivable for the inner range, whereas underwater acoustic communication is considered for the outer range. Under this setup, each robot receives ID codes transmitted from neighboring vehicles within these ranges.

Centered at robot $i$, the communication ranges are defined as two concentric circular zones:

- Inner range: A region within radius $R_{inner}$ from robot $i$, representing the coverage of short-range communication.

- Outer range: A region within radius $R_{outer}$ from robot $i$ ($R_{inner}$ < $R_{outer}$), representing the coverage of relatively long-range communication.

In real-world underwater environments, signal dropouts (packet loss) frequently occur due to multipath propagation, ambient noise, and communication congestion. To emulate this in simulations, a probabilistic model is introduced where signal reception is evaluated as successful only with a specified communication success rate, $P_{comm}$ , even when other robots are within transmission range. Consequently, the number of received signals at each time step fluctuates probabilistically according to $P_{comm}$, meaning that a signal may be received in one time step but missed in another even if robots remain in identical positions. This simplified communication loss captures the macro-level impact of packet dropouts and collisions in underwater acoustic environments.

Specifically, reception success is evaluated with probability $P_{comm}$ for each transmitter-receiver pair at each time step. The reception outcome for a given pair is applied jointly to both inner and outer neighborhood counts; thus, losses of the two signal types are not evaluated independently. Consequently, when a signal from a robot located within the inner range is successfully received, the count is incremented by one for both the inner and outer neighborhoods. Furthermore, reception success in the reverse direction is evaluated independently.

### 2.3 Definition of Potential Function and State Evaluation

At each time step, each robot counts the successfully received signals (ID codes), denoting the counts from the inner and outer neighborhoods as $N_{inner}$ and $N_{outer}$, respectively. Using these two scalar quantities, the potential value $P_i$ which represents the robot's "satisfaction of belonging to the swarm" or "fitness to the swarm" at its current position, is calculated as follows:

$$P_i = N_{inner} - w N_{outer} \quad (1)$$

This formulation is identical to that of Ishida's pattern formation model [19].

Here, w is a morphological parameter (weighting coefficient) that adjusts the relative contributions of the inner and outer neighborhoods. This evaluation function establishes a virtual potential field in which $P_i$ decreases to zero in an isolated state with no surrounding vehicles and increases as the robot approaches the center of the swarm. Consequently, robots can form a swarm by moving toward regions of higher potential $P_i$, although $P_i$ does not necessarily reach its maximum at the geometric center of the swarm. Furthermore, moving in the composite direction of the vector pointing toward higher potential and the external command vector enables swarm formation without relying on conventional methods such as the Boids model.

However, directly using the raw received signal count for action determination could cause the potential value to drop sharply under temporary communication dropouts, potentially triggering excessive transitions to exploratory behavior. In particular, if the number of received signals fortuitously drops at a given time step while the robot is inside the swarm, it may erroneously infer that it has left the swarm and execute an unnecessary large turn. Such hypersensitive responses to instantaneous communication dropouts degrade the stability of both swarm formation and swarm maintenance.

To address this issue, the present study introduces short-term temporal memory to the potential value calculated from the received signal counts, in addition to implementing the communication success rate model. Each robot maintains past potential values as an exponential moving average alongside the instantaneous potential value, using this time-averaged potential to evaluate its state and execute directional changes. This mechanism prevents temporary communication dropouts from immediately triggering large-scale exploratory maneuvers, thereby enhancing swarm stability under imperfect communication conditions.

The potential value calculated by Eq. (1) represents an instantaneous value based on the signals successfully received at each time step, which is hereafter redefined as $P_{raw,i}(t)$:

$$P_{raw,i}(t) = N_{inner,i}(t) - w\,N_{outer,i}(t) \qquad (2).$$

In this study, each robot maintains an exponential moving average of the potential value as an internal state, and the time-averaged potential value $\bar{P}_i(t)$ is updated according to the following equation:

$$\bar{P}_i(t) = (1-\alpha)\bar{P}_i(t-1) + \alpha P_{\mathrm{raw},i}(t) \qquad (3)$$

$\bar{P}_i(t)$ Smoothed potential value at time step t (initialized to zero).

$\bar{P}_i(t-1)$ : Smoothed potential value at the previous time step.

$P_{raw,i}(t)$ : Instantaneous potential value computed from current received signal counts.

$\alpha$ : Time-averaging coefficient ($0<\alpha\leq 1$, with a baseline value of 0.2).

This time-averaged potential value $\bar{P}_i(t)$ is used by each robot to evaluate whether its current position is within the swarm or in an exploratory state. Here, $\alpha$ is the time-averaging coefficient of the potential value, satisfying $0<\alpha\leq 1$. When $\alpha$ is large, the robot responds sensitively to the current reception status; conversely, when $\alpha$ is small, the influence of past states persists strongly, moderating the response to temporary communication dropouts. Considering the trade-off between robustness against communication dropouts and the rate of swarm formation, $\alpha$ = 0.2 is adopted as the baseline value in these simulations, with the effects of varying this parameter presented in Appendix 3-7.

### 2.4 Autonomous Behavioral Decision-Making Algorithm

Each robot determines its heading based on the smoothed potential value $\bar{P}_i(t)$ obtained from Eq. (3), its temporal difference, internal states, and whether an external directional command is received. The following subsections describe the procedures for state evaluation, potential-based directional search, command tracking, and position updating. Through this algorithm, each robot naturally converges toward regions of high potential through probabilistic directional changes without requiring complex relative coordinate calculations via communication, thereby self-organizing and maintaining a swarm.

#### 2.4.1 Determination of Internal States Based on Potential

Each robot possesses two internal states, namely the "stable state" and the "exploratory state," with the initial state set to the exploratory state. In the stable state, the robot determines that its current position is within the swarm and maintains forward motion with high straightness; conversely, in the exploratory state, the robot determines that it has strayed from the swarm and randomly changes its heading to search for regions yielding higher $\bar{P}_i(t)$ values. (Note that the stable state here refers strictly to an internal state within the control algorithm and does not directly guarantee dynamical stability or geometric membership in the swarm.)

- Transition from exploratory to stable state:

$$\overline{P}_i(t) \geq P_{\text{enter}}$$

- Transition from stable to exploratory state:

$$\overline{P}_i(t) < P_{\text{exit}}$$

If neither transition condition is satisfied, the robot maintains its current state. The baseline parameters are set as follows:

$$P_{\text{enter}} = 1.5, \quad P_{\text{exit}} = 1.0$$

Consequently, the previous state is retained within the following range:

$$1.0 \leq \overline{P}_i(t) < 1.5$$

This design implements hysteresis by using distinct thresholds for state entry and exit, thereby preventing frequent state chattering caused by signal fluctuations near the boundaries. Furthermore, a potential threshold $P_{\text{thresh}}$ is defined as a benchmark for swarm stabilization, which is set to the same value as $P_{\text{enter}}$ in the present model.

### 2.4.2 Temporal Potential Difference and Drop Counter

Because underwater robots cannot directly measure the spatial potential gradient despite knowing their own potential values, the gradient is estimated at each update step using the temporal difference of the smoothed potential value, $\Delta\overline{P}_i(t)$, defined as follows:

$$\Delta\overline{P}_i(t) = \overline{P}_i(t) - \overline{P}_i(t-1)$$

where $\overline{P}_i(t)$ and $\overline{P}_i(t-1)$ denote the smoothed potential values at the current and previous time steps, respectively. In addition, each robot maintains a counter tracking the number of potential drops. The counter is initialized to zero and updated according to the following rules:

- If $\Delta\overline{P}_i(t) < 0$ : Increment the counter by one.
- If $\Delta\overline{P}_i(t) > 0$ : Reset the counter to zero.
- If $\Delta\overline{P}_i(t) = 0$ : Leave the counter unchanged.

This update process is executed in both the stable and exploratory states. Because the counter value is retained when the temporal difference is zero, it does not strictly count only drops occurring in consecutive time steps. Furthermore, the counter is reset to zero whenever a large-angle turn, described later, is executed.

Estimating the gradient solely from temporal potential variations $\Delta\overline{P}_i(t)$ provides a practical approach for information-constrained underwater robots, operating on the same principle as bacterial chemotaxis. Generally, determining the spatial potential gradient $\nabla P$ requires simultaneous measurements across multiple points around the vehicle or sophisticated multi-element receiver arrays. In contrast, by using the temporal difference resulting from vehicle motion as the evaluation metric, the proposed method achieves directional decision-making using only a single omnidirectional receiver (hydrophone) and minimal microcontroller computation.

### 2.4.3 Heading Determination in the Exploratory State

In the exploratory state, each robot determines its search direction based on the sign of $\Delta\overline{P}_i(t)$ and the drop counter. Let $\boldsymbol{V}_{swarm}$ denote the unit vector representing this search direction.

- If $\Delta\overline{P}_i(t) > 0$ : Maintain the current heading as the search direction.
- If $\Delta\overline{P}_i(t) < 0$ : A left or right turn is chosen with equal probability (0.5 each), and the turning angle is determined according to the drop count. In the baseline configuration, the turning angle is drawn uniformly from 0° to 60° when the updated drop count is 1 or 2; if the count reaches 3 or more, the angle is chosen uniformly between 90° and 180°, after which the counter is reset to zero. The orientation resulting from this turn becomes the search direction.
- If $\Delta\overline{P}_i(t) = 0$ : A uniform random perturbation between -10° and +10° is added to the current heading, and this updated orientation becomes the search direction.

This method selects whether to maintain or alter the heading based on temporal variations in the robot's own potential value. Through this process, robots avoid large heading alterations in response to minor potential drops caused by temporary communication dropouts, transitioning to full-scale exploratory behavior only when potential decreases persist. This mechanism temporally distinguishes instantaneous fluctuations in received signal counts from genuine departures from the swarm. Furthermore, because potential values are influenced not only by ego-motion but also by the displacement of neighboring robots and communication reception outcomes, $\boldsymbol{V}_{swarm}$ does not always align with the ascending gradient of the potential field.

Directional search in this model is formulated not as a simple random walk, but as a robust search algorithm that integrates time-averaged potential values, hysteresis-based state retention, and stepwise directional adjustments driven by persistent potential drops.

### 2.4.4 External Directional Command Tracking and Vector Composition

To perform tasks such as wide-area surveys, a target travel direction (angle $\theta_{\mathrm{cmd}}$) is assumed to be transmitted from an external source (such as a support vessel) to all robots. Each robot is assumed to determine its heading relative to a common directional reference using an onboard compass. The reception of the command at each update step varies probabilistically according to the command arrival probability $P_{\mathrm{cmd}}$. Let $V_{\mathrm{cmd}}$ denote the unit vector representing the command direction.

In the stable state, the robot adopts $\theta_{\mathrm{cmd}}$ as its target direction if a command is received, while maintaining its current heading if no command is received. Under this state, vector composition using the weight $W_{\mathrm{mix}}$, described later, is not applied.

In the exploratory state, the handling of the command depends on whether a turn triggered by a potential drop is executed. When a turn is executed because $\Delta\overline{P}_i(t) < 0$, the search direction resulting from the turn is adopted as the target direction without performing vector composition, regardless of whether a command is received.

Conversely, when a turn is not executed (i.e., $\Delta\overline{P}_i(t) \geq 0$) and a command is successfully received, $\boldsymbol{V}_{swarm}$ and $\boldsymbol{V}_{cmd}$ are combined using the following equation:

$$\boldsymbol{V}_{next} = (1 - W_{\text{mix}})\,\boldsymbol{V}_{swarm} + W_{\text{mix}}\,\boldsymbol{V}_{cmd} \qquad (4)$$

where $W_{\text{mix}}$ is a weighting coefficient for the command direction, with the baseline configuration set as follows:

$$W_{\text{mix}} = 0.2$$

The directional angle of the composite vector $\boldsymbol{V}_{next}$ defines the target direction, whereas its magnitude is used solely for directional evaluation and is not applied to the translation speed. If no command is received under this condition, the direction of $\boldsymbol{V}_{swarm}$ is adopted as the target direction.

**2.4.5 Update of Heading and Position**

When executing a turn triggered by a potential drop, the current heading is directly changed to the post-turn target direction chosen randomly. Otherwise, the shortest angular difference from the current heading to the target direction is computed, and the heading is adjusted by 10% of this difference according to the following equation:

$$\theta_{\text{updated}} = \theta_{\text{current}} + 0.1\Delta\theta_{\text{shortest}}$$

where the shortest angular difference $\theta_{\text{shortest}}$ is defined as a signed difference that accounts for angular periodicity. Furthermore, an angular perturbation uniformly distributed between −2° and +2° is added to the updated heading, regardless of the internal state or whether a turn was executed:

$$\theta_{\text{final}} = \theta_{\text{updated}} + \theta_{\text{noise}}$$

The translation speed of each robot is set to $v_{\text{mean}}$, and the positional coordinates $(x, y)$ are updated according to the final heading. Periodic boundary conditions are applied to the computational domain, such that individuals moving outside the boundary re-enter from the opposite side. The next state of each individual is computed based on the identical pre-update configuration, and all individuals are updated synchronously.

This model serves as a proof-of-concept model for planar motion in two dimensions and does not incorporate hydrodynamic forces, vehicle inertia, or turning performance constraints.

Furthermore, to isolate and clarify the fundamental effectiveness of the algorithm, this model does not incorporate repulsive behaviors for collision avoidance upon close approach, such as those in the Boids model. Collision avoidance algorithms remain a subject for future investigation. Possible approaches include proximity avoidance based on the received signal strength from neighboring vehicles. Additionally, because potential values reach local maxima and gradients flatten near the center of a swarm, collision avoidance could be achieved by prioritizing command-tracking behavior along the command vector in such regions.

**2.5 Simulation Implementation**

The proposed model was implemented using HTML and JavaScript, enabling it to run directly on standard web browsers. All simulation parameters can be configured through a user interface (UI), ensuring that all results presented in this paper are fully reproducible by modifying these settings.

The computational domain is defined as an 800 × 800 pixel two-dimensional space, where one pixel represents the unit length of the model. Each robot is rendered with a visual length of 14 pixels, which serves for graphical display rather than a physical dimension for collision detection. Velocities are expressed in model length units (pixels) per time step, without a direct mapping to physical time or distance.

The parameters configurable via the graphical UI are as follows:

- Random seed value
- Maximum number of steps (the time step at which the simulation terminates and exports results)
- Number of robots
- Mean speed ($v_{\mathrm{mean}}$)
- Command direction ($\theta_{\mathrm{cmd}}$)
- Inner neighborhood range ($R_{inner}$)
- Outer neighborhood range ($R_{outer}$)
- Morphological parameter ($w$)
- Mixing weight for command tracking ($W_{mix}$)
- Potential threshold ($P_{thresh}$, equivalent to $P_{enter}$)
- Command arrival probability ($P_{cmd}$)
- Communication success rate ($P_{comm}$)

Upon reaching the maximum number of steps, the following time-series data are exported as a CSV file:

- Time step count
- Total number of swarms
- Overall mean potential value (average of smoothed potential values $P_i(t)$ across all individuals)
- Cluster-level metrics for each swarm (radius, individual count, density, mean potential P, and mean orientation vector)

Furthermore, the population size and initial conditions were configured as follows:

- Number of robots: Set to 80 vehicles in the baseline case.
- Initial placement: Coordinates and headings (ranging from −180° to 180°) are generated randomly. The initial and running speeds of all robots are fixed to the common value $v_{\mathrm{mean}}$, and each vehicle is assigned a unique ID.
- Boundary conditions: Periodic boundary conditions are applied to position updates, inter-vehicle distance calculations, and swarm connectivity evaluations.

- Update method: The subsequent state of each individual is computed from the current configuration, and all vehicle positions are updated synchronously.

Random numbers are generated using the Xorshift algorithm [22], a pseudo-random number generator proposed by George Marsaglia in 2003 that is characterized by high computational efficiency and straightforward implementation.

### 2.6 Definition of Quantitative Metrics

#### 2.6.1 Fundamental Metrics for Quantitative Evaluation

To quantify robot swarms, the simulation results were evaluated primarily based on the following metrics:

1. Total number of swarms
2. Number of individuals belonging to swarms
3. Mean radius and density of each swarm
4. Overall mean potential value, $\bar{P}$

To identify individuals belonging to a swarm, the stable state evaluation based on the time-averaged potential value and hysteresis described in Section 2.4 was employed. Individuals identified in the stable state were selected as candidates for swarm membership, and their spatial coordinates were mapped onto a 40 × 40 grid composed of square cells with a side length of 20 pixels. Subsequently, connected component labeling using eight-neighbor connectivity was applied to identify spatially connected clusters of individuals as distinct swarms. The total number of swarms and the number of member individuals were determined based on these labeling results. Periodic boundary conditions were also applied during connectivity evaluation. The details of this algorithm are provided in Appendix 2.

The radius of each swarm was calculated from the spatial distribution of the robots after identifying individual clusters using the detection algorithm, and the swarm density was subsequently computed from the member count and the calculated radius.

The overall mean potential value was calculated as the average of smoothed potential values across all robots including non-member individuals (i.e., the sum of smoothed potential values divided by the total number of robots), whereas the mean potential value of each swarm was computed exclusively for its constituent members. These metrics reflect local states based on received signal counts, where higher values indicate that robots are concentrated within the pattern regions (high-potential areas) of the Ishida model. However, an increase in potential alone does not unequivocally imply convergence to a steady state or an improvement in communication efficiency.

#### 2.6.2 Quantitative Metrics for Evaluating Swarm Formation Success

To evaluate simulation results objectively under noisy conditions, three quantitative metrics are introduced:

**1. Swarm Participation Rate ($R_{\mathrm{part}}$)**

- **Definition:** The proportion of robots successfully incorporated into swarms. Specifically, it represents the percentage of all robots classified into the stable state according to Section 2.4 and subsequently detected as swarm members.
- **Range:** 0.0 (0%) to 1.0 (100%)
- **Calculation:**

$$R_{\text{part}} = \frac{N_{\text{stable}}}{N_{\text{total}}} \times 100\ [\%]$$

where $N_{\text{stable}}$ is the number of robots in the stable state, and $N_{\text{total}}$ is the total number of robots.

**2. Polarization (Alignment Parameter, $\Phi_{align}$)**

- **Definition:** A metric indicating the degree of directional alignment and collective motion across the swarm, widely used in flocking studies and Boids models.
- **Range:** 0.0 (completely disordered) to 1.0 (perfectly aligned in an identical direction)
- **Calculation:** Defined as the magnitude of the sum of heading unit vectors across all swarm-affiliated robots, divided by the total number of affiliated robots:

$$N_{\text{affiliated}} = \sum_k N_k$$

$$S_X = \sum_k S_{X,k}, \quad S_Y = \sum_k S_{Y,k}$$

$$\Phi_{\text{align}} = \frac{\sqrt{{S_X}^2 + {S_Y}^2}}{N_{\text{affiliated}}}$$

where $N_k$ denotes the number of robots in the $k$-th swarm, and $S_X$ and $S_Y$ represent the total sums of the directional vector components across all swarms.

**3. Command Tracking Score ($S_{\text{track}}$)**

- **Definition:** A metric evaluating the degree of alignment between the robot travel directions and the external command direction.
- **Range:** −1.0 (moving directly opposite to the command) to 1.0 (perfect alignment with the command)
- **Calculation:** The average inner product between the heading unit vector of each swarm-affiliated robot and the command direction unit vector:

$$S_{\text{track}} = \frac{S_X V_{cmd,x} + S_Y V_{cmd,y}}{N_{\text{affiliated}}}$$

where $V_{cmd,x}$ and $V_{cmd,y}$ are the X and Y components of the unit command vector $\boldsymbol{V}_{\text{cmd}}$.

## 3. Results

### 3.1 Baseline Parameter Configuration and Simulation Results

Table 1 lists the baseline parameter configuration. Because this model involves numerous parameters resulting in a vast combinatorial space with an unknown emergence domain, a set of conditions under which swarm formation was consistently observed during preliminary exploration was selected as the baseline parameter set. Each parameter was varied individually from these baseline conditions to evaluate changes in swarm formation behavior. Simulations were performed using a random seed of 12345 as the baseline; therefore, this baseline configuration does not necessarily represent the global optimum across the entire parameter space.

**Table 1.** Baseline parameter configuration.

| Parameter | Value |
|---|---|
| Random seed | 12345 |
| Number of robots | 80 |
| Mean speed ($v_{mean}$) | 1.2 |
| Outer neighborhood range ($R_{outer}$) | 120 |
| Inner neighborhood range ($R_{inner}$) | 60 |
| Morphological parameter (w) | 0.4 |
| Command angle ($\theta_{cmd}$) | 0° |
| Mixing weight ($W_{mix}$) | 0.2 |
| Command arrival probability ($P_{cmd}$) | 1 |
| Communication success rate ($P_{comm}$) | 1 |
| Potential time-averaging coefficient (α) | 0.2 |
| Potential threshold ($P_{thresh}$=$P_{enter}$) | 1.5 |
| Stable state exit threshold ($P_{exit}$) | 1 |
| Consecutive drop threshold for large turn ($n_{drop}$) | 3 |
| Standard directional adjustment coefficient | 0.1 |
| Angular perturbation for all updates | −2° to +2° |
| Search direction perturbation for zero difference | −10° to +10° |
| Small-angle turning range | 0° to 60° |
| Large-angle turning range | 90° to 180° |

Figure 2 illustrates the simulation evolution from Step 0 (initial randomly scattered state) to Step 2000 under the baseline parameter set. The prominent arrow in the simulation window indicates the directional command (Command Angle = 0°), while the color of each robot shifts toward red as its local potential value increases. The baseline simulation results reveal that robots aggregate into high-potential regions over time, demonstrating that swarm formation can be achieved without relying on the Boids model or requiring visual and absolute positional data.

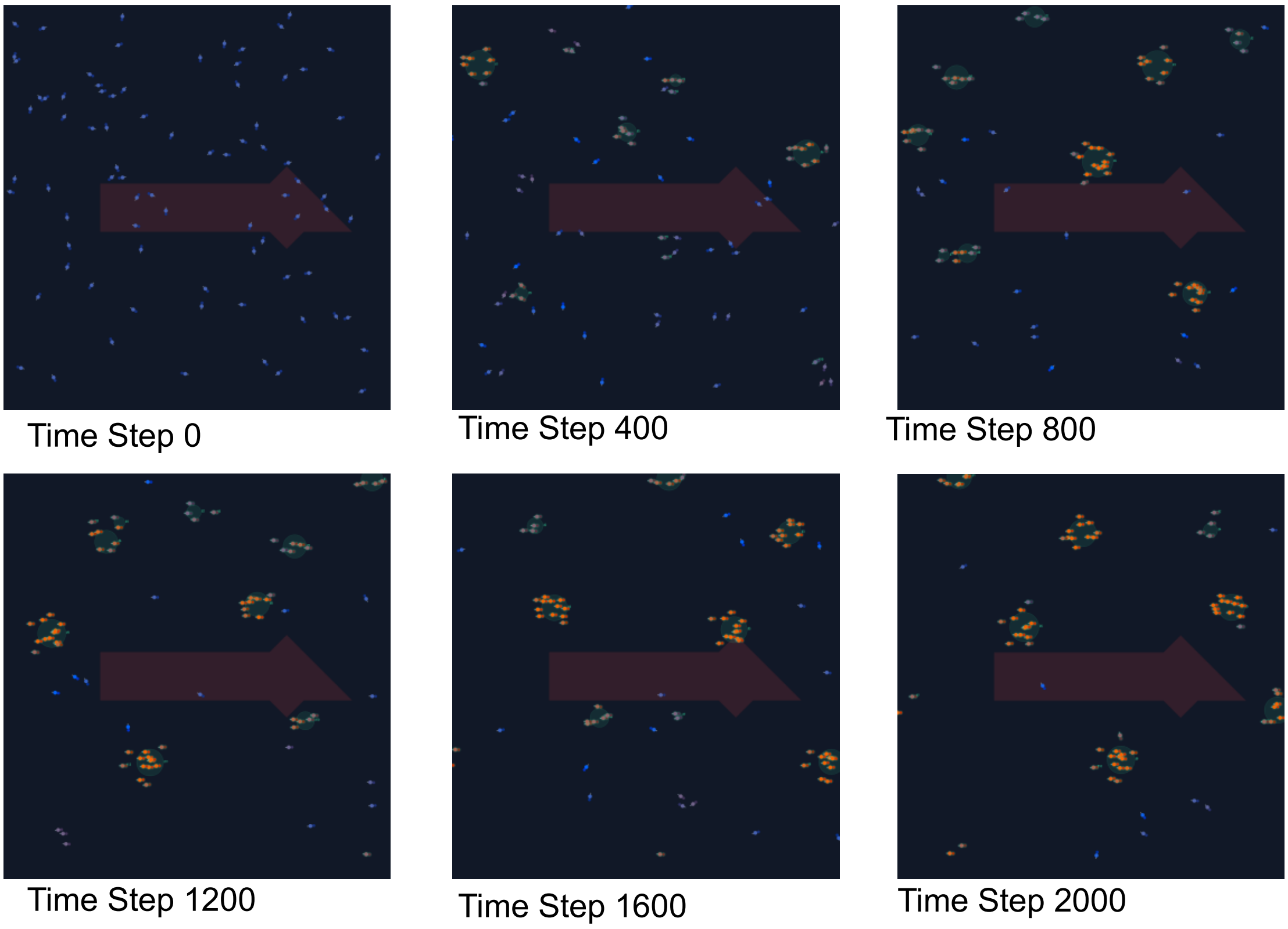

**Figure 2.** Snapshots of the simulation time series under the baseline parameter set, showing the evolution from Step 0 (initial randomly scattered state) to Step 2000. The prominent arrow indicates the directional command (Command Angle = 0°). The color of each robot transitions toward red in regions of higher potential values.

Next, to quantitatively evaluate the autonomous swarm formation process from a randomized initial configuration, Figure 3 presents the time-series profiles of the total swarm count, affiliated individual count, mean potential value $\bar{P}$, swarm participation rate, command tracking score, and polarization across time steps. As time elapses, the overall mean potential value and the swarm participation rate increase steadily, while the number of swarms converges to a constant value following initial formation fluctuations; concurrently, both the command tracking score and polarization remain stably near 1.0. These findings quantitatively demonstrate that, even in the absence of centralized control, local rule-based interactions among individuals drive the macroscopic system toward coherent swarm formation characterized by elevated potential values and stabilized cluster counts.

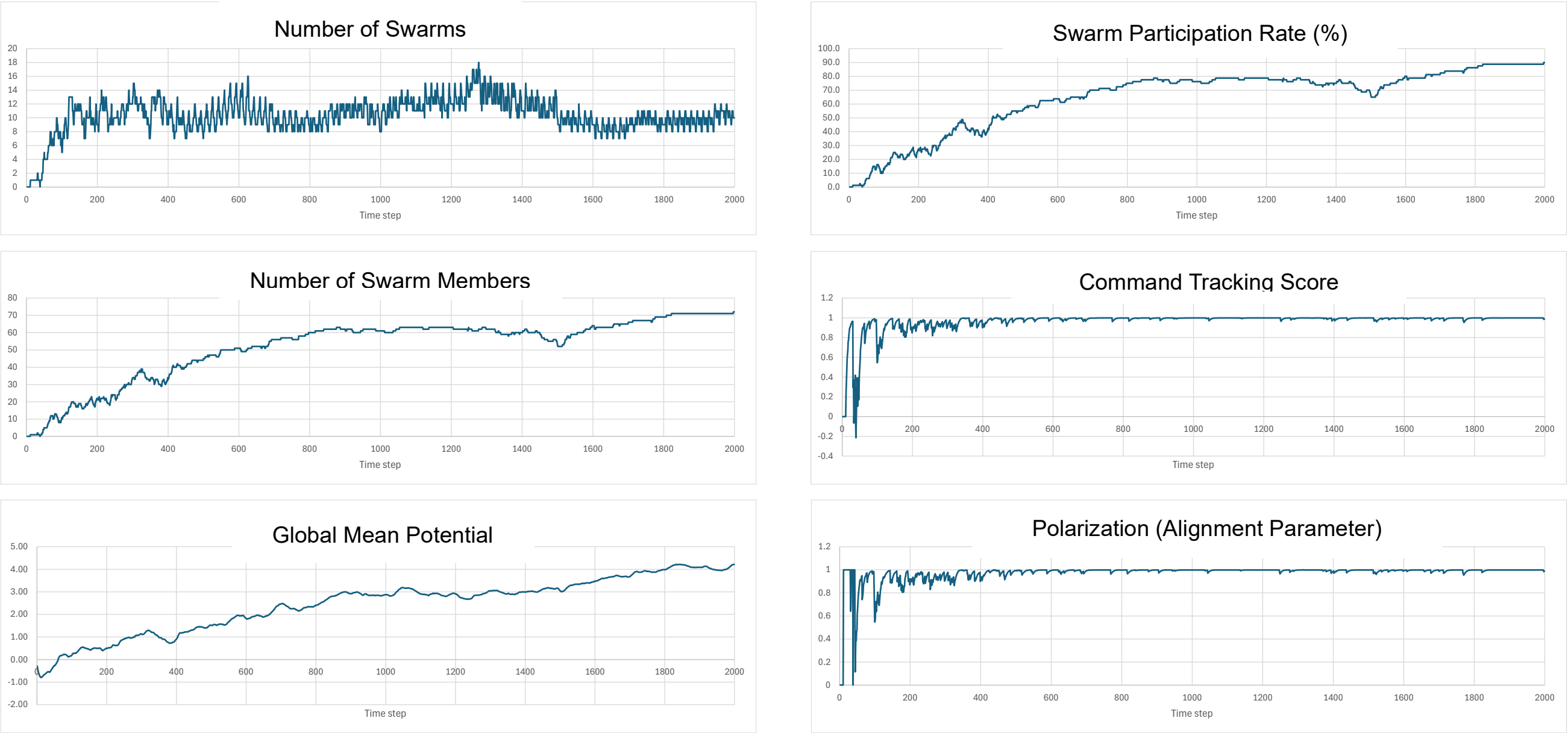

**Figure 3.** Time-series evolution of total swarm count, affiliated individuals, mean potential value $\bar{P}$, swarm participation rate, command tracking score, and polarization under the baseline parameter set. The overall mean potential and participation rate increase steadily over time, while the number of swarms stabilizes after transient fluctuations, and both the command tracking score and polarization remain close to 1.0.

Next, to evaluate model robustness (i.e., the capacity to sustain stable swarm formation unaffected by stochastic variations), Table 2 presents the results obtained by varying the random seed under the baseline parameter configuration. Five cases were tested by varying the seed by up to $\pm 200$ relative to the baseline seed of 12345, and the metrics evaluated at the final step (Step 2000) were compiled. Although microscopic stochastic variations emerge in the total number of formed swarms (ranging from 7 to 12) and their sizes due to localized density fluctuations from randomized initial placements, polarization exceeds 0.98 across all cases, and the swarm participation rate remains stable at a high level between 86% and 94%. This demonstrates that the proposed algorithm can reliably generate macroscopic cooperative behaviors (collective locomotion and group cohesion) without being governed by initial spatial contingencies.

**Table 2.** Comparison of evaluation metrics across five random seeds under the baseline parameter set (seed 12345 represents the baseline case).

| Seed | w | Communication Success Rate | Command Success Rate | Number of Swarms | Number of Swarm Members | Mean Swarm Radius | Mean Swarm Density | Global Mean Potential ($\bar{P}$) | Swarm Participation Rate (%) | Command Tracking Score | Polarization (Alignment Parameter) |
|---|---|---|---|---|---|---|---|---|---|---|---|
| 12345 | 0.4 | 1 | 1 | 10 | 72 | 18.1 | 0.099 | 4.21 | 90.0 | 0.9993 | 0.9939 |
| 12145 | 0.4 | 1 | 1 | 7 | 69 | 21.3 | 0.050 | 5.07 | 86.3 | 0.9999 | 0.9991 |
| 12245 | 0.4 | 1 | 1 | 12 | 73 | 15.1 | 0.111 | 4.39 | 91.3 | 0.9998 | 0.9994 |
| 12445 | 0.4 | 1 | 1 | 11 | 75 | 18.2 | 0.035 | 4.73 | 93.8 | 0.9999 | 0.9992 |
| 12545 | 0.4 | 1 | 1 | 9 | 74 | 18.6 | 0.096 | 5.17 | 92.5 | 0.9998 | 0.9987 |

### 3.2 Parametric Control of Swarm Formation

To demonstrate the versatility of the proposed model, simulations were conducted by varying individual parameters relative to the baseline configuration. This investigation aimed to determine how varying parameters modulates swarm characteristics, rather than merely evaluating whether swarm formation is feasible.

#### 3.2.1 Effect of Morphological Parameter $w$

Figure 4 presents the evaluation metrics at the terminal state (Step 2000) when the morphological parameter $w$ is varied from 0.2 to 0.8, including total swarm count, mean potential value $\bar{P}$, command tracking score, polarization, mean swarm density, mean swarm radius, and swarm participation rate. Figure 5 displays the spatial distributions of the swarms at the terminal state (Step 2000). Figure 6 shows the time-series evolution of the mean potential value $\bar{P}$ and swarm participation rate under various values of the morphological parameter $w$.

These results indicate that swarm formation is hindered when $w \geq 0.60$, with no swarms emerging at $w = 0.80$; conversely, swarms form stably when $w \leq 0.40$, where decreasing $w$ reduces cluster size and increases the total number of swarms. Notably, the mean swarm radius peaks at the transitional value of $w = 0.60$.

These findings suggest the practical applicability of the proposed model to distinct maritime missions. For instance, setting $w \approx 0.40$ establishes dispersed swarms suitable for broad-scale seabed bathymetric mapping, whereas lowering $w$ forms tightly packed clusters ideal for localized, intensive inspections of underwater structures.

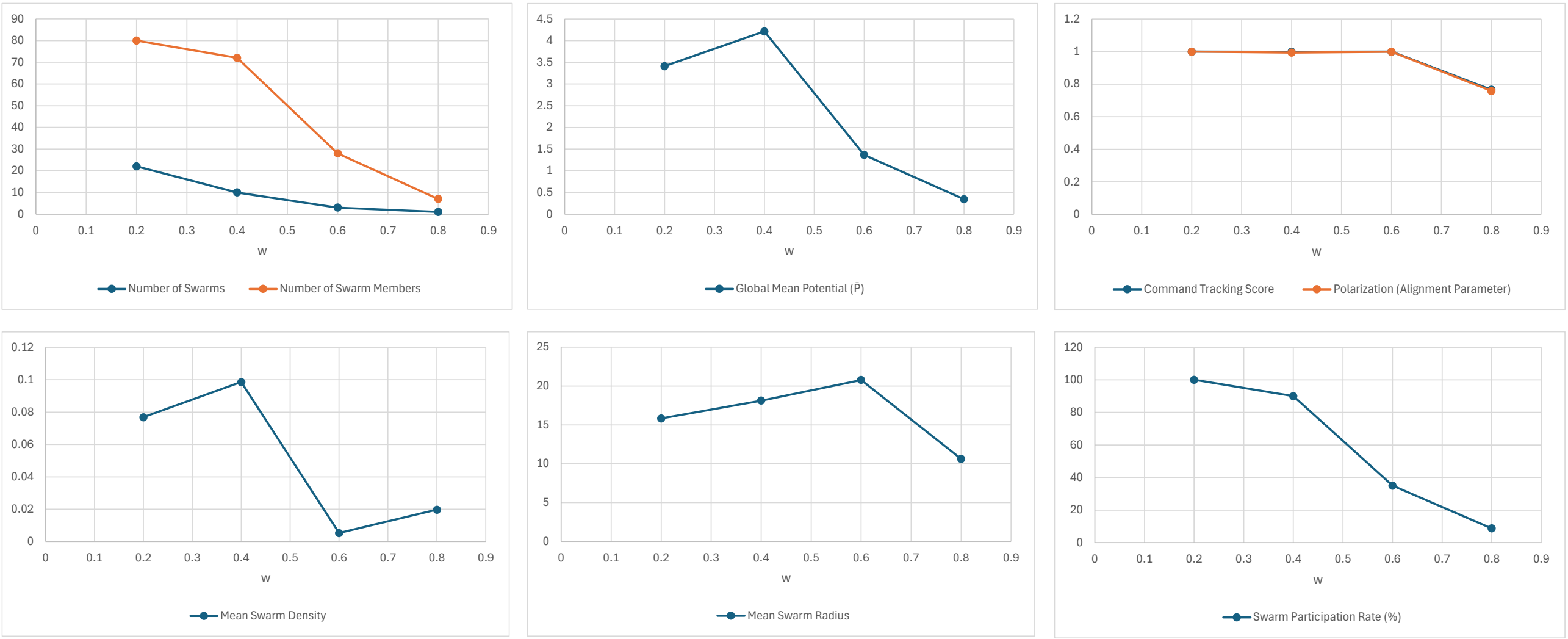

**Figure 4.** Evaluation metrics at the terminal state (Step 2000) under varying morphological parameters $w$, including total swarm count, mean potential value $\bar{P}$, command tracking score, polarization, mean swarm density,

mean swarm radius, and swarm participation rate. Swarm formation is suppressed when $w \geq 0.60$ and fails completely at $w = 0.80$. In contrast, swarms emerge successfully when $w \leq 0.40$, where decreasing $w$ leads to smaller individual clusters and an increased number of swarms. The mean swarm radius peaks at $w = 0.60$.

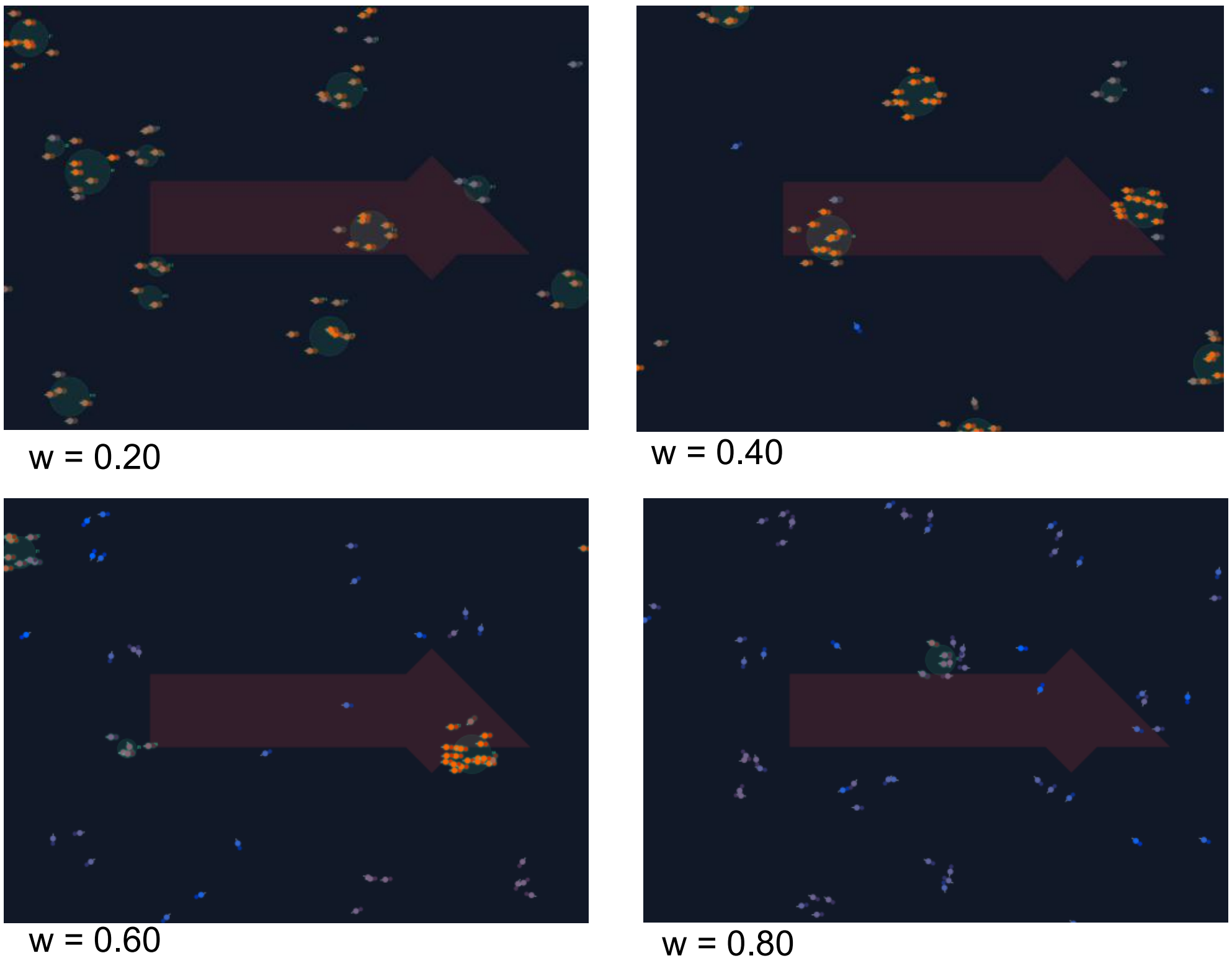

**Figure 5.** Spatial distributions of swarms at the terminal state (Step 2000) under different morphological parameters $w$. Swarm formation is impaired when $w \geq 0.60$ and is completely absent at $w = 0.80$. Conversely, swarms form when $w \leq 0.40$, with smaller cluster sizes and a higher number of swarms observed as $w$ decreases.

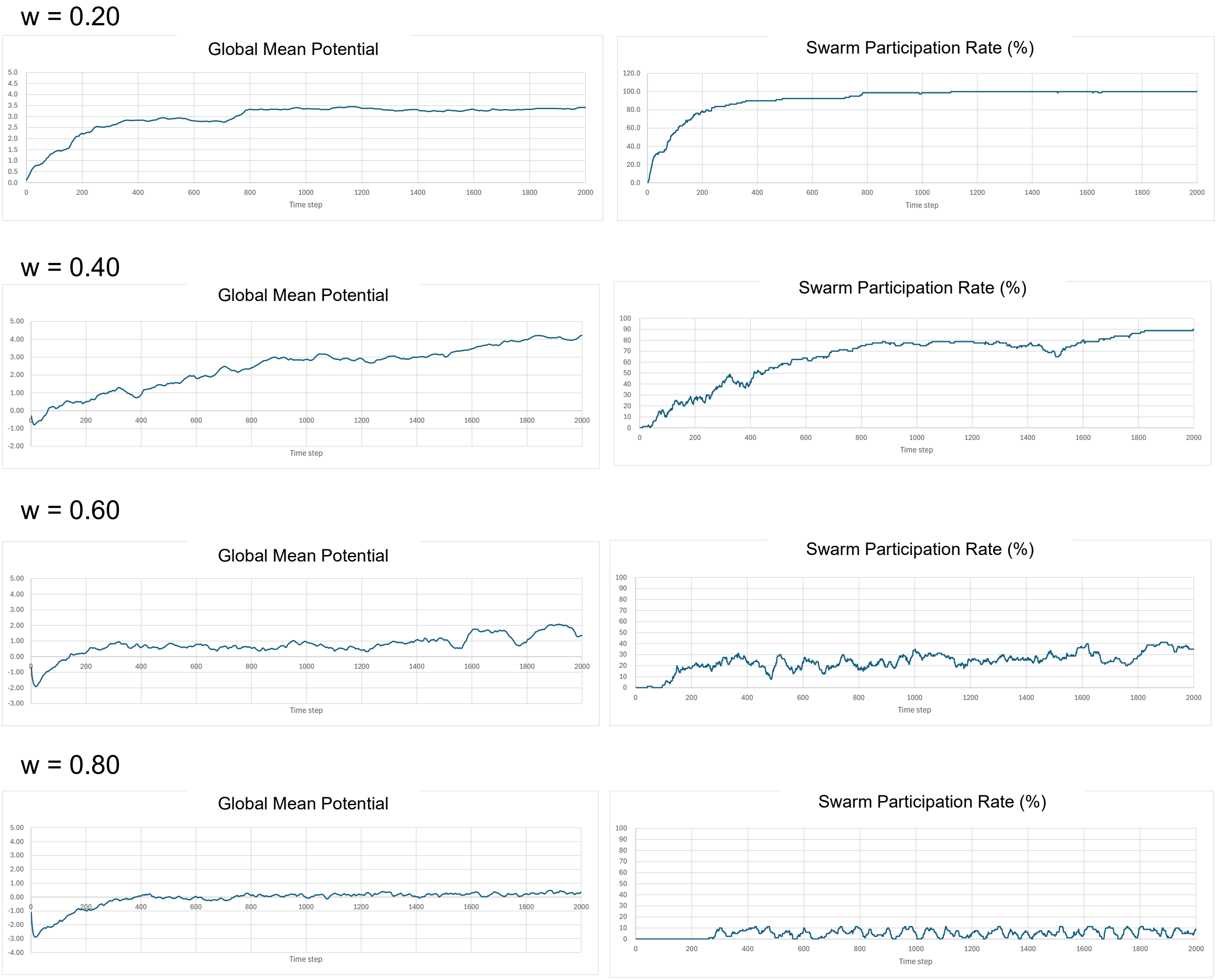

**Figure 6.** Time-series evolution of evaluation metrics including mean potential value $\bar{P}$ and swarm participation rate under various values of $w$. When $w > 0.60$, the mean potential value approaches zero, accompanied by a sharp decline in the swarm participation rate.

### 3.2.2 Effect of Potential Threshold $P_{\text{thresh}}$

In the proposed model, moving toward regions of higher potential $P$ corresponds to an intrinsic motivation to remain within the swarm, whereas following the directional command corresponds to an extrinsic motivation to reach a destination. Here, the potential threshold $P_{\text{thresh}}$ (equivalent to $P_{\text{enter}}$) serves as the criterion for determining whether a robot is located within the swarm, and this investigation examines how its setting influences swarm cohesion and mobility.

**Table 3.** Evaluation metrics at Step 2000 under various potential threshold settings $P_{\text{thresh}}$ (random seed fixed at 12345, $P_{\text{exit}} = 1.0$).

| P Threshold | Seed | w | Communication Success Rate | Command Success Rate | Number of Swarms | Number of Swarm Members | Mean Swarm Radius | Mean Swarm Density | Global Mean Potential ($\bar{P}$) | Swarm Participation Rate (%) | Command Tracking Score | Polarization (Alignment Parameter) |
|---|---|---|---|---|---|---|---|---|---|---|---|---|
| 1.00 | 12345 | 0.4 | 1 | 1 | 13 | 70 | 14.3 | 0.104 | 3.69 | 87.5 | 0.9999 | 0.9997 |
| 1.25 | 12345 | 0.4 | 1 | 1 | 9 | 69 | 20.9 | 0.040 | 4.15 | 86.3 | 0.9998 | 0.9992 |
| 1.50 | 12345 | 0.4 | 1 | 1 | 10 | 72 | 18.1 | 0.099 | 4.21 | 90.0 | 0.9993 | 0.9939 |
| 1.75 | 12345 | 0.4 | 1 | 1 | 7 | 65 | 21.7 | 0.051 | 3.82 | 81.3 | 0.9999 | 0.9993 |
| 2.00 | 12345 | 0.4 | 1 | 1 | 3 | 54 | 29.8 | 0.006 | 5.71 | 67.5 | 0.9999 | 0.9991 |
| 2.25 | 12345 | 0.4 | 1 | 1 | 6 | 62 | 25.9 | 0.005 | 4.31 | 77.5 | 0.9999 | 0.9990 |
| 2.50 | 12345 | 0.4 | 1 | 1 | 0 | 0 | 0.0 | 0.000 | 1.36 | 0.0 | 0.0000 | 0.0000 |

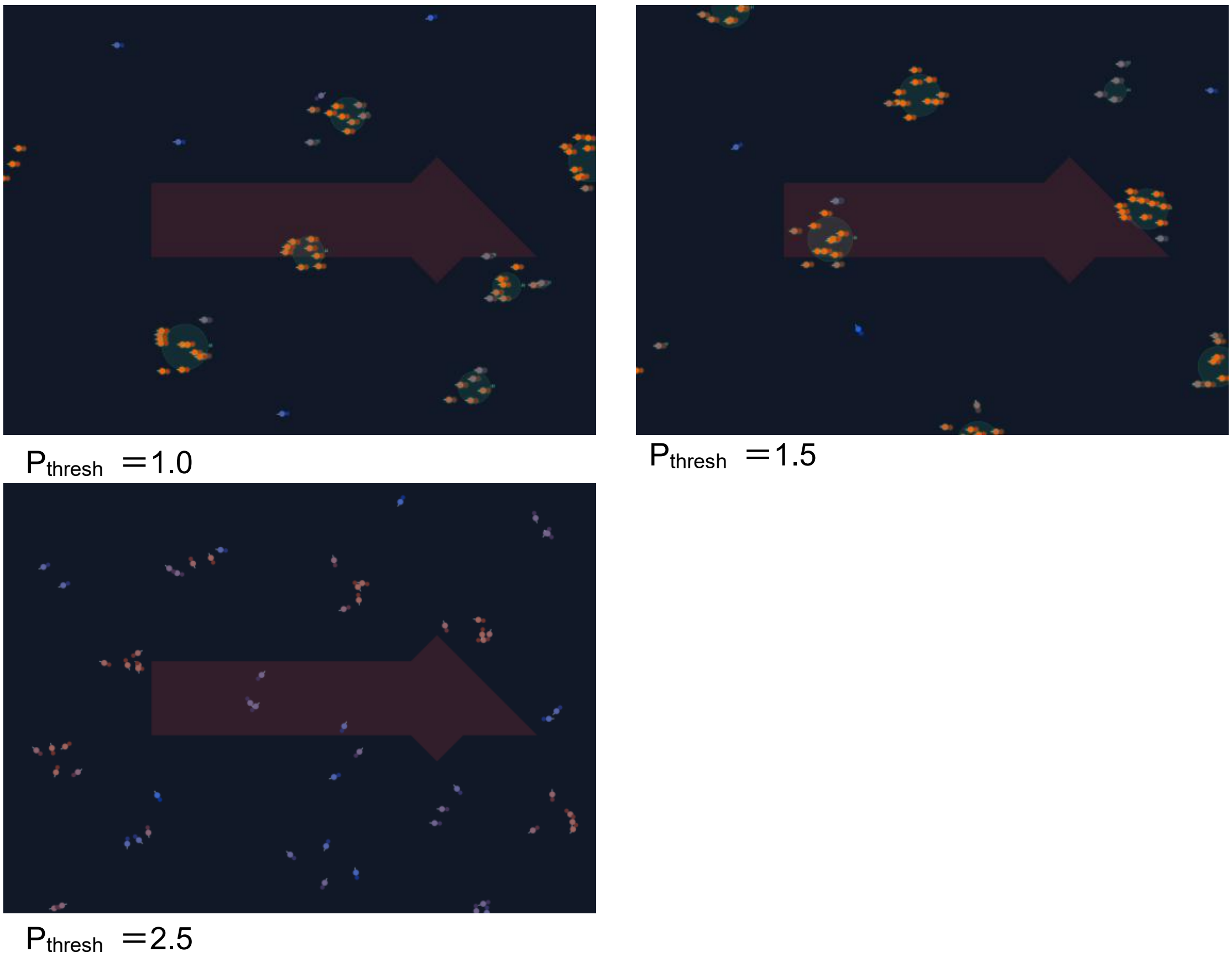

**Figure 7.** Swarm formation results at the terminal state (Step 2000) under different potential thresholds $P_{\text{thresh}}$, showing that swarms cease to form when $P_{\text{thresh}}$ reaches 2.50.

### 3.2.3 Sensitivity Analysis of Other Parameters

Evaluation metrics at the terminal state (Step 2000) obtained by varying parameters other than those described above are summarized in Appendix 3.

## 3.3 Robustness to Communication Dropouts

To verify model robustness against communication dropouts, which is a critical consideration for practical deployment in real marine environments, simulations were conducted under varying command arrival probabilities and communication success rates.

### 3.3.1 Effect of Command Arrival Probability (Command Success Rate)

Figure 8 and Table 4 present the quantitative evaluation metrics, including the mean potential value $\bar{P}$ and total swarm count, when the command arrival probability ($P_{\mathrm{cmd}}$) is varied from 0.1 to 1.0 for random seed 12345. Additionally, Figure 9 illustrates the spatial distributions of robots at the terminal state (Step 2000) under varying command arrival probabilities.

Furthermore, Figure 10 summarizes the mean and standard deviation of the mean potential value $\bar{P}$ and swarm participation rate evaluated across five random seeds (12145, 12245, 12345, 12445, and 12545) under various command arrival probabilities. These results confirm that high potential values and swarm participation rates are consistently maintained regardless of the random seed.

These findings demonstrate that swarms emerge successfully even when the command arrival probability drops to approximately 0.5, verifying that the proposed model possesses sufficient robustness against command loss.

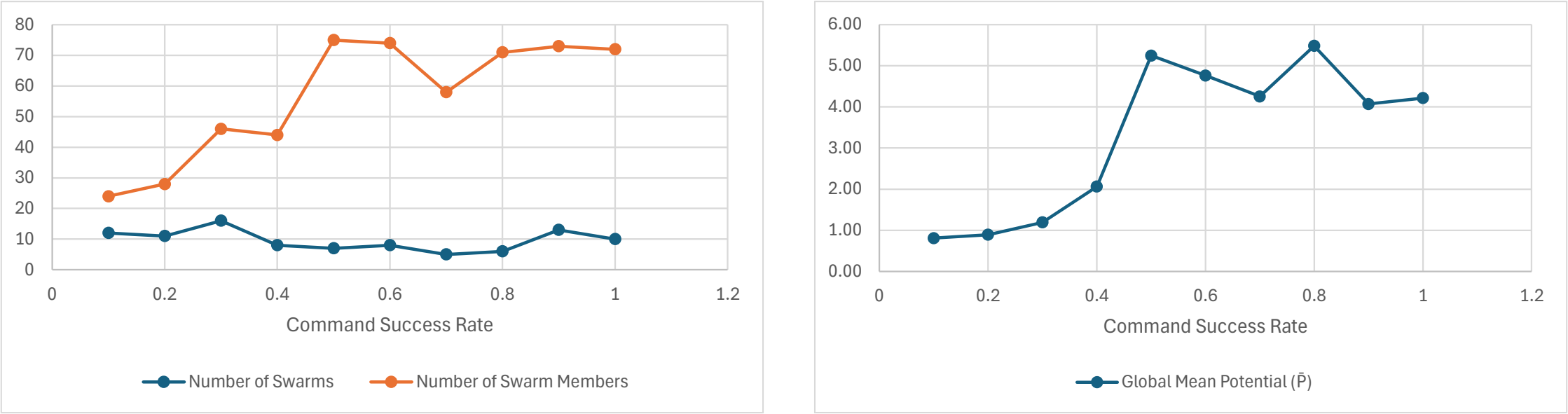

**Figure 8.** Total swarm count, number of affiliated individuals, and mean potential value $\bar{P}$ under varying command arrival probabilities. Swarm formation is confirmed and high mean potential values are maintained even at a command arrival probability of 0.5.

**Table 4.** Evaluation metrics under varying command arrival probabilities (for random seed 12345).

| Command Success Rate | Seed | w | Communication Success Rate | Command Success Rate | Number of Swarms | Number of Swarm Members | Mean Swarm Radius | Mean Swarm Density | Global Mean Potential (P̄) | Swarm Participation Rate (%) | Command Tracking Score | Polarization (Alignment Parameter) |
|---|---|---|---|---|---|---|---|---|---|---|---|---|
| 0.1 | 12345 | 0.4 | 1 | 0.1 | 12 | 24 | 6.3 | 0.190 | 0.81 | 30.0 | 0.6735 | 0.8348 |
| 0.2 | 12345 | 0.4 | 1 | 0.2 | 11 | 28 | 10.3 | 0.123 | 0.89 | 35.0 | 0.6884 | 0.8022 |
| 0.3 | 12345 | 0.4 | 1 | 0.3 | 16 | 46 | 8.0 | 0.129 | 1.19 | 57.5 | 0.9566 | 0.9403 |
| 0.4 | 12345 | 0.4 | 1 | 0.4 | 8 | 44 | 14.5 | 0.122 | 2.06 | 55.0 | 0.9679 | 0.9586 |
| 0.5 | 12345 | 0.4 | 1 | 0.5 | 7 | 75 | 23.4 | 0.050 | 5.24 | 93.8 | 0.9999 | 0.9982 |
| 0.6 | 12345 | 0.4 | 1 | 0.6 | 8 | 74 | 23.5 | 0.005 | 4.76 | 92.5 | 0.9999 | 0.9984 |
| 0.7 | 12345 | 0.4 | 1 | 0.7 | 5 | 58 | 26.7 | 0.005 | 4.25 | 72.5 | 0.9999 | 0.9983 |
| 0.8 | 12345 | 0.4 | 1 | 0.8 | 6 | 71 | 23.8 | 0.058 | 5.48 | 88.8 | 0.9997 | 0.9972 |
| 0.9 | 12345 | 0.4 | 1 | 0.9 | 13 | 73 | 16.2 | 0.077 | 4.07 | 91.3 | 0.9998 | 0.9993 |
| 1.0 | 12345 | 0.4 | 1 | 1 | 10 | 72 | 18.1 | 0.099 | 4.21 | 90.0 | 0.9993 | 0.9939 |

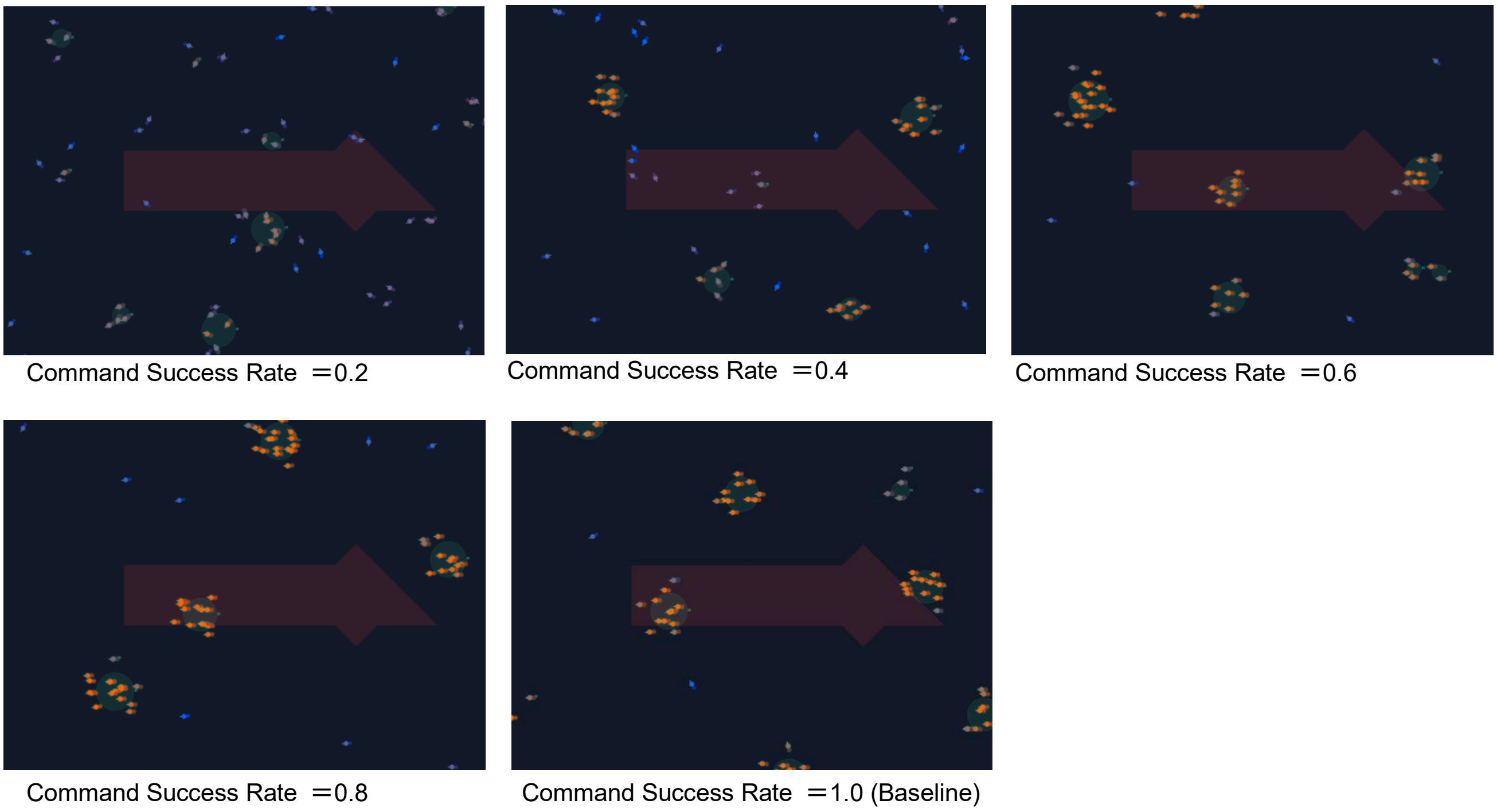

**Figure 9.** Spatial distributions of robots at the terminal state (Step 2000) under varying command arrival probabilities (for random seed 12345), confirming that swarms form when the command arrival probability is 0.5 or higher.

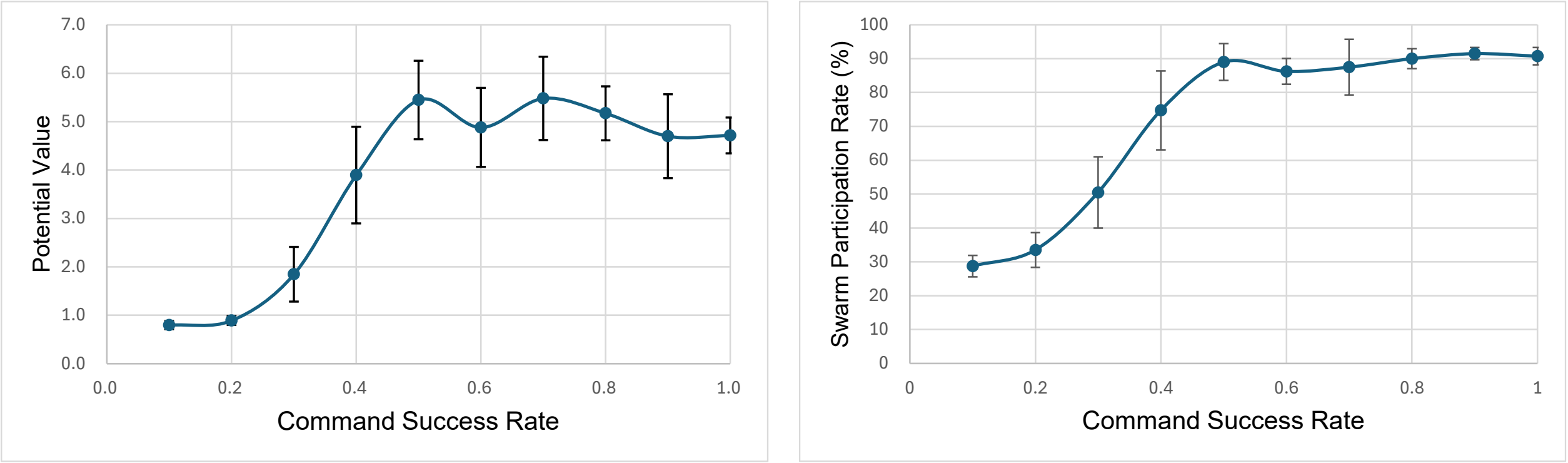

**Figure 10.** Mean potential value $\bar{P}$ and swarm participation rate under varying command arrival probabilities (mean $\pm$ standard deviation across five random seeds), showing that high potential values and participation rates are maintained regardless of the random seed.

### 3.3.2 Effect of Communication Success Rate

Figure 11 and Table 5 present the quantitative evaluation metrics, including total swarm count and mean potential value $\bar{P}$, when the communication success rate ($P_{\text{comm}}$) is varied from 0.6 to 1.0. Figure 12 illustrates the spatial distributions of robots at the terminal state (Step 2000) under different communication success rates. For baseline seed 12345, the terminal swarm participation rates were 90.0%, 82.5%, 67.5%, 57.5%, and 18.8% for communication success rates of 1.0, 0.9, 0.8, 0.7, and 0.6, respectively; although multiple swarms formed even at $P_{\text{comm}} = 0.7$, the participation rate degraded noticeably compared with ideal communication.

Furthermore, Figure 13 summarizes the mean and standard deviation of the mean potential value $\bar{P}$ and swarm participation rate across five random seeds (12145, 12245, 12345, 12445, and 12545), demonstrating that the participation rate consistently exceeds 50% once $P_{\text{comm}}$ reaches approximately 0.7 or higher regardless of stochastic initialization.

These findings indicate that swarms can be autonomously formed and maintained as long as the communication success rate is approximately 0.7 or higher (even with a 30% packet loss), confirming that the proposed model retains a viable swarm formation capability under moderate communication loss.

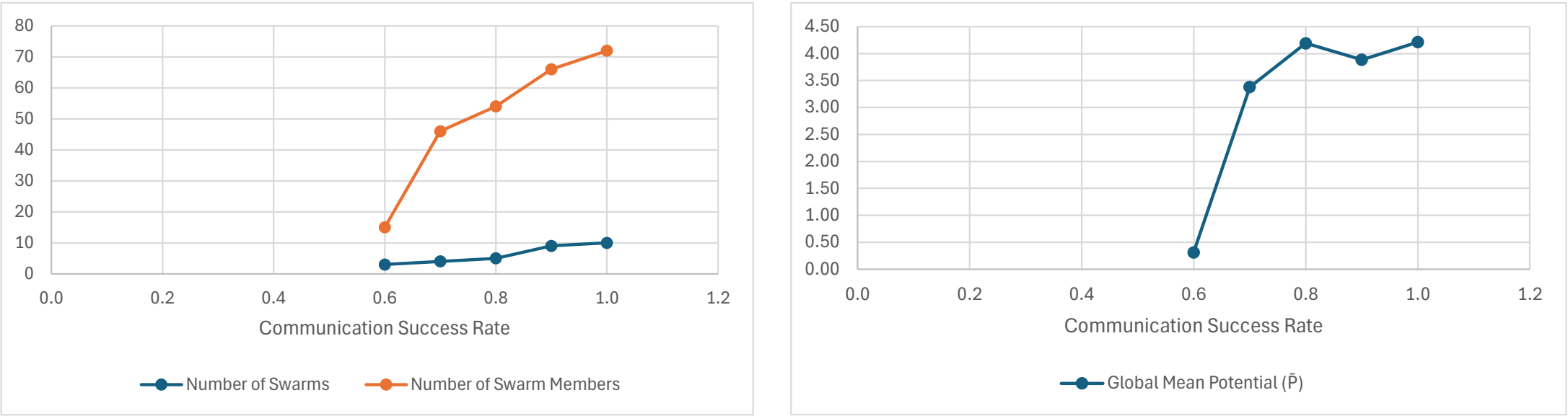

**Figure 11.** Total swarm count, affiliated individual count, and mean potential value $\bar{P}$ under varying communication success rates. Multiple swarms emerge and the mean potential value increases sharply when the communication success rate exceeds 0.7.

**Table 5.** Evaluation metrics under varying communication success rates.

| Communication Success Rate | Seed | w | Communication Success Rate | Command Success Rate | Number of Swarms | Number of Swarm Members | Mean Swarm Radius | Mean Swarm Density | Global Mean Potential (P̄) | Swarm Participation Rate (%) | Command Tracking Score | Polarization (Alignment Parameter) |
|---|---|---|---|---|---|---|---|---|---|---|---|---|
| 0.6 | 12345 | 0.4 | 0.6 | 1 | 3 | 15 | 17.4 | 0.006 | 0.31 | 18.8 | 0.990 | 0.998 |
| 0.7 | 12345 | 0.4 | 0.7 | 1 | 4 | 46 | 18.9 | 0.084 | 3.38 | 57.5 | 0.998 | 0.996 |
| 0.8 | 12345 | 0.4 | 0.8 | 1 | 5 | 54 | 17.5 | 0.074 | 4.19 | 67.5 | 0.994 | 0.997 |
| 0.9 | 12345 | 0.4 | 0.9 | 1 | 9 | 66 | 17.7 | 0.075 | 3.89 | 82.5 | 0.996 | 0.978 |
| 1.0 | 12345 | 0.4 | 1 | 1 | 10 | 72 | 18.1 | 0.099 | 4.21 | 90.0 | 0.999 | 0.994 |

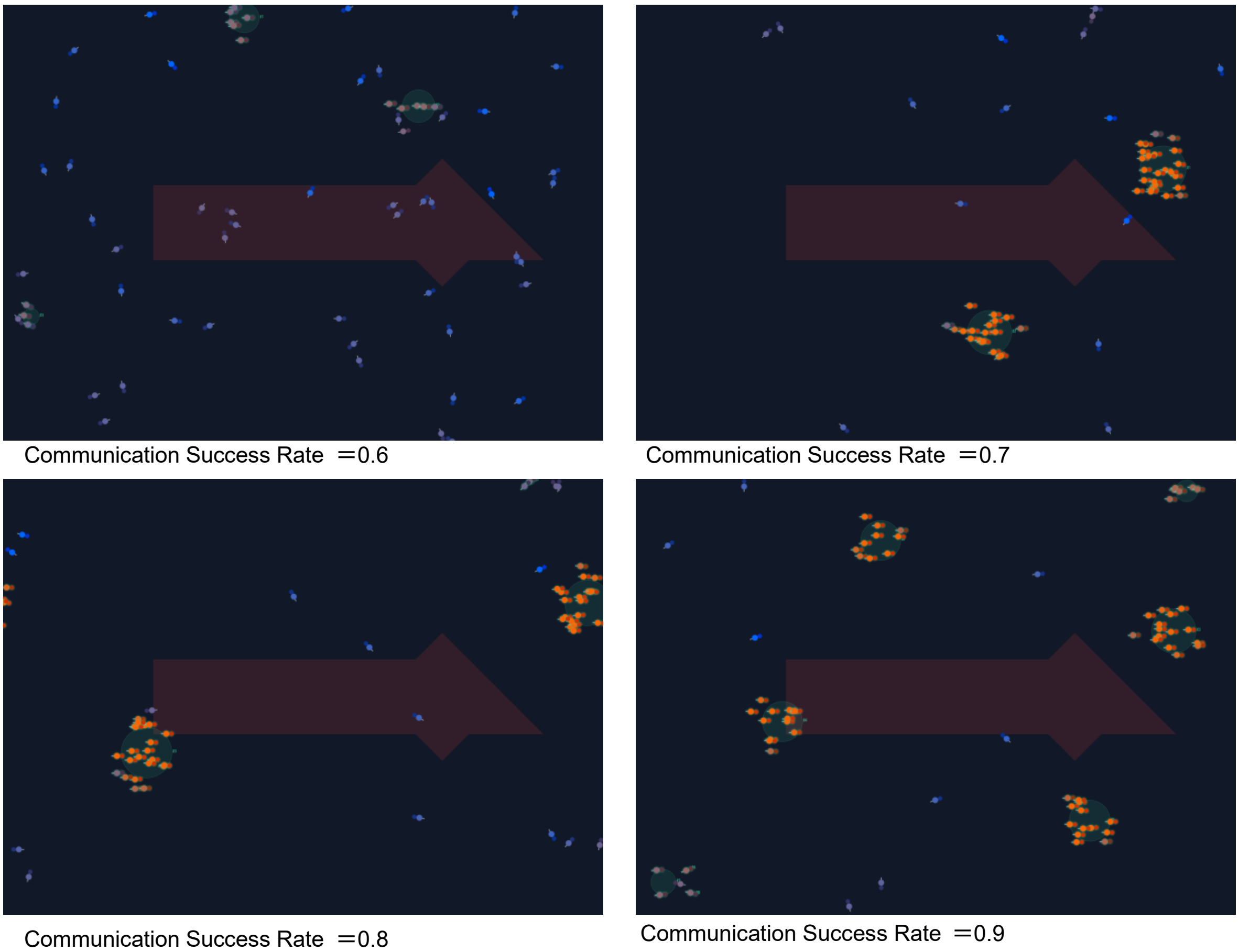

**Figure 12.** Spatial distributions of robots at the terminal state (Step 2000) under varying communication success rates. Swarm formation is not observed at a communication success rate of 0.6, whereas multiple swarms are formed when the rate exceeds 0.7.

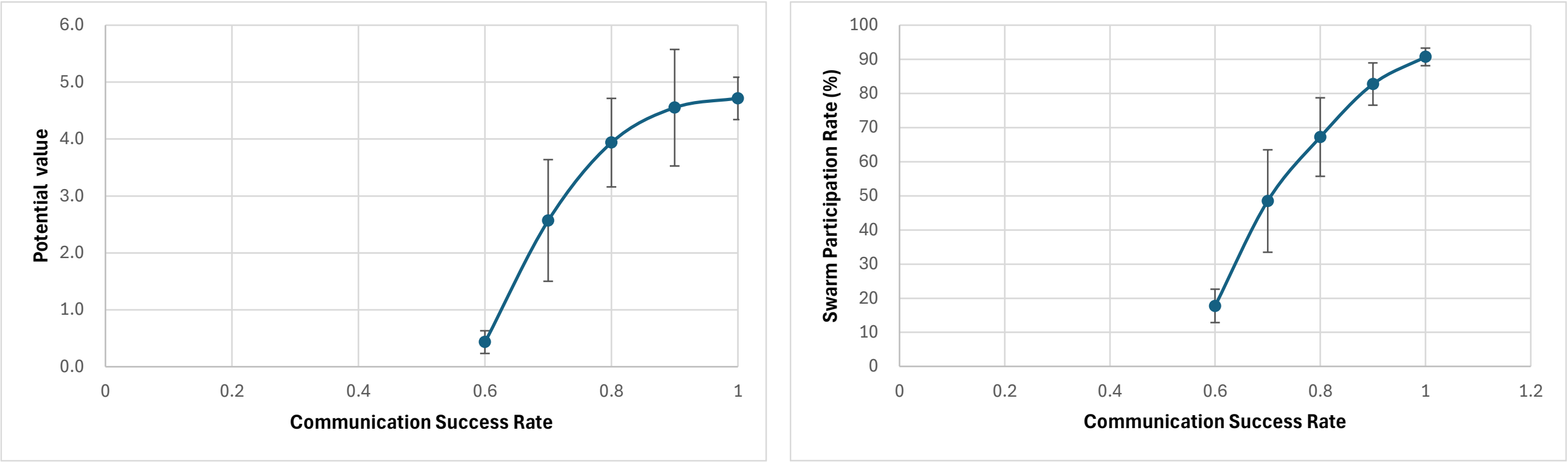

**Figure 13.** Mean potential value $\bar{P}$ and swarm participation rate under varying communication success rates (mean $\pm$ standard deviation across five random seeds), showing that the swarm participation rate consistently exceeds 50% when the communication success rate is approximately 0.7 or higher regardless of the random seed.

## 4. Discussion

This study demonstrated that the self-organized formation of a robot swarm in continuous space can be achieved using two types of communication signals with different transmission ranges, where each robot counts only the number of signals received within its inner and outer neighborhoods. Under the baseline parameter set, 80 randomly deployed robots formed multiple swarms over time, with the overall mean potential value $P$, swarm participation rate, and polarization converging to stable values. Even when the random seed was varied, the swarm participation rate remained generally within the range of 86% to 94%, and polarization maintained high values. These results suggest that the proposed model does not depend strongly on specific initial configurations or stochastic random-walk contingencies, thereby achieving swarm formation with consistent reproducibility.

A key feature of the proposed model is that it extends the self-organizing pattern formation model of the Ishida model, which is based on counting neighborhood states, to the control of mobile robot swarms. In the Ishida model, stable spatial patterns such as spots or stripes emerge from the balance between the state counts in the inner and outer neighborhoods, and in the present study, the potential $P$ calculated by each robot corresponds to a locally stable region within this pattern formation process. Robots probabilistically perform exploratory behavior toward high-potential regions formed in their vicinity; consequently, the robot swarm as a whole forms multiple localized aggregation clusters. Thus, rather than predefining the center position or geometric formation of the swarm, the proposed model self-organizes the spatial structure of the swarm solely from local counts of communication signals exchanged among robots.

### 4.1 Controllability of Swarm Behavior

#### 4.1.1 Control via the Morphological Parameter w

The morphological parameter w adjusts the relative weighting between the received signal counts from the inner and outer neighborhoods, serving as a primary control parameter governing the spatial structure of swarms in the proposed model. Simulation results demonstrated that varying w altered the total swarm count, mean swarm radius, and swarm participation rate.

When w is small, the inhibitory effect from the outer neighborhood is relatively weak, allowing robots to stabilize readily with nearby neighbors; consequently, multiple small clusters tend to emerge, increasing the total number of swarms. Conversely, as w increases, the influence of the outer neighborhood strengthens, causing the potential P to decrease more readily due to neighboring robots distributed across a wider area, which impedes entry into the stable state and makes distinct swarm formation difficult at $w = 0.8$.

Furthermore, the mean swarm radius tended to increase around $w = 0.6$, which is attributed to the stronger inhibitory influence of the outer neighborhood preventing excessive aggregation and leading to more spatially extended clusters. However, as w increases further, local high-potential regions required to sustain cohesion become difficult to establish, rendering swarm formation itself unstable.

These results indicate that tuning w enables control over the spatial dispersion of the robot swarm. For instance, setting w to a moderate value can generate multiple moderately dispersed swarms suitable for broad-area surveys of seabed topography or environmental concentration distributions; conversely, setting w to a lower value can produce more localized, high-density clusters for intensive inspections of specific structures or anomalies. However, because excessively large values of w cause swarm formation to break down, practical applications require selecting a task-specific optimal value primarily within the range of $w = 0.2$ to 0.6.

#### 4.1.2 Effects of Individual Parameters on Swarm Formation

Parameters other than $w$ were also confirmed to influence the feasibility of swarm formation, cluster size, swarm participation rate, and directional stability.

First, the results obtained by varying the number of robots (Appendix 3-1) suggest that swarm formation requires a population density above a certain threshold. When the robot population was 40, few individuals were incorporated into swarms, making stable swarm formation difficult; in contrast, for populations of 80 or more robots, the swarm participation rate increased substantially, with the majority of robots joining swarms at populations of 100 or 120. This indicates that because the proposed model relies on the number of signals received within the communication range, an excessively low robot density prevents individual vehicles from acquiring sufficient local information, thereby hindering the establishment of a well-defined potential field.

Regarding the mean speed, excessively low speeds restricted the spatial search range, tending to impede the progression of swarm formation (Appendix 3-2); conversely, increasing the mean speed enabled robots to explore a broader area, facilitating access to high-potential regions and increasing the overall mean potential value $P$. However, because excessively high speeds could cause vehicles to overshoot swarms or fail to execute timely turns, future physical implementations require further investigation incorporating turning performance and vehicle inertia.

Regarding the communication ranges (Appendices 3-3 and 3-4), increasing the outer neighborhood range tended to improve both the swarm participation rate and the mean potential value $P$, which is attributed to each robot being able to detect neighboring vehicles over a wider area and perform potential evaluations reflecting the macroscopic spatial structure of the entire swarm. Conversely, when the inner neighborhood range was enlarged, the swarm participation rate improved while the total swarm count also increased, which is attributed to the expanded inner neighborhood allowing more robots to readily achieve locally high $P$ values and form multiple stable regions. Therefore, the ratio between the inner and outer neighborhood ranges serves as a critical parameter governing whether the robots organize into a few large clusters or disperse into numerous small clusters.

Regarding the potential threshold $P_{\text{thresh}}$ (Section 3.2.2), this parameter plays a critical role as both the criterion for swarm membership and the condition governing transitions between the exploratory and stable states. When $P_{\text{thresh}}$ is low, robots enter the stable state relatively easily, resulting in a high swarm participation rate accompanied by a tendency to form numerous small clusters. Conversely, when $P_{\text{thresh}}$ is high, robots transition into the stable state only in regions with sufficiently high potential values, reducing the number of swarm-affiliated individuals; in particular, no swarm formation was observed at $P_{\text{thresh}} = 2.5$. This finding indicates that an excessively high $P_{\text{thresh}}$ prevents robots located inside a cluster from being classified into the stable state, causing them to persist in exploratory behavior. Therefore, while $P_{\text{thresh}}$ enhances swarm cohesion, setting this threshold excessively high inhibits swarm formation.

Regarding the mixing weight $W_{\text{mix}}$ (Appendix 3-6), this parameter governs the balance between tracking external directional commands and maintaining swarm cohesion. Specifically, when a robot is in the exploratory state, receives a command, and does not execute a turn triggered by a potential drop, $W_{\text{mix}}$ defines the composition ratio between the search direction and the command direction, whereas it is not applied to command tracking in the stable state. In Table A3-6 of Appendix 3, the swarm participation rates for $W_{\text{mix}}$ values of 0.1, 0.2, 0.3, and 0.4 were 100.0%, 90.0%, 73.8%, and 81.3%, respectively, exhibiting a non-monotonic variation. Although the command tracking score remained high across all settings, notable differences appeared in the swarm participation rate. Setting a lower mixing weight tended to increase the swarm participation rate by prioritizing swarm maintenance; conversely, increasing the mixing weight strengthened movement toward the command direction, potentially reducing swarm cohesion slightly. Therefore, rather than simply amplifying the command vector, an appropriate weighting between intrinsic swarm-forming behavior and extrinsic command-tracking behavior is essential.

The effect of the time-averaging coefficient $\alpha$, which governs the smoothness of the response to communication dropouts, is presented in Appendix 3-7. When $\alpha$ is small, the influence of past potential values persists strongly, preventing overreaction to temporary communication dropouts; conversely, when $\alpha$ is large, the robot responds sensitively to the current reception status, accelerating adaptation to environmental changes at the cost of increased susceptibility to communication noise. Under conditions of a communication success rate of 1.0 and random seed 12345, the swarm participation rates for $\alpha = 0.1$, 0.2, 0.3, and 0.4 were 86.3%, 90.0%, 87.5%, and 87.5%, respectively. Within this range, the participation rate peaked at $\alpha = 0.2$, whereas the mean potential value $P$ reached its maximum at $\alpha = 0.1$. Consequently, the baseline setting of approximately $\alpha = 0.2$ appears to offer a relatively favorable balance between robustness against communication dropouts and swarm formation speed in terms of both swarm participation rate and mean potential $P$, although further investigation of this parameter is warranted.

## 4.2 Comparison of Required Information for Swarm Control

A key characteristic of the proposed model is that each robot requires substantially less information than in the conventional Boids model, a representative swarm formation framework.

In the Boids model, each individual executes three behavioral rules with respect to surrounding individuals (separation, alignment, and cohesion), which requires acquiring the relative positions, distances, bearings, and, in some cases, velocity vectors of neighboring individuals. Even in two-dimensional space, at least relative coordinates or distance and bearing information is required for each neighboring individual. Furthermore, performing alignment behavior additionally requires the headings or velocity vectors of neighboring vehicles. In a three-dimensional underwater environment, the addition of depth information further increases the volume of data that each vehicle must acquire and process.

In contrast, in the proposed model, the information required by each robot consists of only two scalar quantities, namely the number of robots within the inner neighborhood ($N_{\mathrm{inner}}$) and that within the outer neighborhood ($N_{\mathrm{outer}}$), eliminating the need to acquire the exact relative coordinates, distances, bearings, or velocity vectors of individual neighboring vehicles. Each robot counts the received ID signals and evaluates its local state using $P = N_{\mathrm{inner}} - wN_{\mathrm{outer}}$. Subsequently, based on the temporal difference between the current and previous $P$ values, the robot decides whether to maintain its current heading or alter its direction.

This distinction is particularly significant in underwater environments, where Global Positioning System (GPS) signals are unavailable, radio communication is severely limited, and acoustic communication suffers from narrow bandwidth, transmission delays, and packet loss. Consequently, it is challenging for individual robots to share the positions and velocities of numerous neighboring vehicles at high frequencies. Because the proposed model reduces communication requirements to broadcasting simple presence-indicating ID signals and counting received signals, it is well suited for marine environments with severely constrained communication bandwidth.

However, the proposed model does not replace all functionalities of the Boids model; in particular, a collision avoidance function corresponding to the separation rule has not yet been explicitly incorporated. This omission is because the present study focuses on verifying a self-organizing swarm formation principle based on dual-neighborhood signal counts. Specifically, this principle demonstrates that clustering and directional guidance emerge solely from counting scalar quantities without computing relative position vectors. Therefore, the results of this study should be interpreted not as achieving complete swarm control encompassing collision avoidance and precise formation keeping, but rather as demonstrating the feasibility of realizing the fundamental functions of swarm formation and collective movement along a command direction using minimal information. For future physical implementations, short-range collision avoidance and obstacle avoidance mechanisms must be integrated while preserving the low-information communication overhead of the proposed model.

### 4.3 Communication Realism and Robustness

In the control of underwater robot swarms, robustness against communication imperfections is a critical evaluation criterion. In the present study, the command arrival probability and the inter-robot communication success rate were varied independently to investigate the impact of communication dropouts on swarm formation.

Regarding the command arrival probability, both the swarm participation rate and polarization maintained high values even when the probability was reduced to 0.6. In the proposed model, each robot has an opportunity to receive the external command at every time step and adjusts its heading toward the command direction upon reception in the stable state. Consequently, even when the command arrival probability decreases, command tracking can be sustained through intermittent direct reception by individual robots. Specifically, the proposed model does not require external commands to reach every robot in complete synchronization; instead, collective movement along the command direction is achieved through local interactions within the swarm, which is particularly advantageous in underwater environments where communication bandwidth and communication success rates are severely constrained.

In contrast, the inter-robot communication success rate directly influenced swarm formation itself. As the communication success rate decreased from 1.0 to 0.9 and 0.8, swarm formation was sustained despite reductions in the swarm participation rate and mean potential value $P$. Furthermore, even at a communication success rate of 0.7, more than half of all robots remained affiliated with swarms, and the formation of multiple clusters was confirmed. This demonstrates that the proposed model retains a viable swarm formation capability even under conditions where approximately 30% of communication signals are lost.

However, when the communication success rate decreased to 0.6, the swarm participation rate dropped sharply, and the mean potential value $P$ also fell to a low level. This result indicates that the proposed model is not infinitely robust against communication dropouts, and that a minimum signal reception density is required to sustain swarm formation. Specifically, while the proposed model eliminates the need for each robot to acquire detailed positional information of neighboring vehicles, it requires a communication success rate sufficient to reliably estimate the received signal counts within the inner and outer neighborhoods. Based on the present results under the baseline parameter configuration, a communication success rate of approximately 0.7 approaches the lower threshold for swarm formation, whereas a rate of 0.8 to 0.9 or higher is desirable to achieve a practically stable swarm participation rate. However, because these findings were obtained under a specific parameter set and simulation duration, further investigation is required to establish general swarm formation limits and communication performance requirements for physical vehicles.

Three underlying mechanisms account for the robustness against communication dropouts achieved in the present study. First, each robot evaluates its internal state using a time-averaged potential value rather than instantaneous received signal counts; consequently, temporary communication dropouts are unlikely to trigger an immediate transition to the exploratory state. Second, because hysteresis is introduced by separating the entry and exit thresholds for the stable state, frequent state switching is prevented even when the potential value $P$ fluctuates near the thresholds. Third, because a large-angle turn is executed only when potential decreases occur consecutively, excessive exploratory behavior in response to sporadic communication dropouts is suppressed.

Through these mechanisms, the proposed model maintains stable macroscopic self-organization across the entire swarm even in the presence of temporary communication dropouts and fluctuations in received signal counts. This demonstrates that even without complete information, individual robots can generate collective order across the entire swarm by acting through the temporal integration of local and incomplete information. Therefore, the proposed model offers potential as a low-information and robust swarm control scheme in environments with severe communication and localization constraints, such as underwater domains. However, the effects described above reflect the intended design mechanisms, and the present simulations evaluated a combined model incorporating all three mechanisms. Therefore, the individual contribution of each mechanism must be verified separately in future work.

Furthermore, the present communication model is a simplified representation that probabilistically reduces signal reception probability. Consequently, it does not explicitly account for physical phenomena observed in actual underwater acoustic communication, such as transmission delays, asynchrony, multipath propagation, distance-dependent attenuation, and packet collisions. Therefore, future work must move beyond modeling the communication success rate as a uniform probability by introducing a channel model in which the success rate varies dynamically with inter-robot distance, relative bearing, swarm density, and the number of simultaneous transmissions. Moreover, toward physical implementation, further validation is required to account for duplicate reception of ID signals, simultaneous reception from multiple vehicles, communication cycle intervals, and reception latency.

## 5. Conclusions

This study proposed a decentralized control model for underwater robot swarms that utilizes received signal counts from dual neighborhoods with different transmission ranges alongside temporal memory. Without relying on the relative positions or velocity vectors of neighboring vehicles as control inputs, each robot determines its behavior based on a smoothed potential value, its internal state, and a common directional command. The simulation results confirmed that the proposed model enables autonomous robot swarm formation through appropriate parameter tuning.

**1) Advantage of the Proposed Model:** Whereas the conventional Boids model requires the relative positions (coordinates) of neighboring vehicles, the proposed model achieves swarm control solely using a scalar quantity representing the number of vehicles within the communication range. This drastic reduction in communication overhead is considered highly advantageous in underwater environments.

**2) Extensibility to Three-Dimensional Space:** Although the proposed model was investigated in a two-dimensional framework, individual AUVs in wide-area marine exploration (such as seabed bathymetric mapping) can readily maintain a constant depth using onboard pressure sensors and depth controllers. Therefore, reducing the problem to cooperative swarm control within a horizontal two-dimensional plane is practically well justified in ocean engineering. Furthermore, the algorithm itself can be readily extended to three dimensions using spherical neighborhood ranges. Accordingly, future work will focus on developing the targeted three-dimensional simulation model of swarm robots operating in a three-dimensional marine environment.

**3) Introduction of Quantitative Evaluation:** Beyond visually confirming that swarms were formed, the present results successfully quantified swarm characteristics such as swarm density, total swarm count, and degree of spatial dispersion, as detailed in the Results and Discussion sections.

Based on the present simulation results, future work must conduct numerical experiments in which underwater robots operate within an underwater environment simulation model, alongside investigations into physical vehicles. For physical implementations, various uncertainties not accounted for in the present model must also be addressed, including compass accuracy, the effects of underwater magnetic disturbances, and noise in communication receivers.

## ACKNOWLEDGMENTS

**CRediT authorship contribution statement:** Takeshi Ishida: Conceptualization, Methodology, Software, Validation, Formal analysis, Investigation, Data curation, Visualization, Writing – original draft, Writing – review and editing, Funding acquisition, Project administration.

**Funding:** This research was supported by grants from the Japan Society for the Promotion of Science, KAKENHI Grant Number 23K04283.

**Competing interests:** The author declares that he has no known competing financial interests or personal relationships that could have appeared to influence the work reported in this paper.

**Preprint statement:** A preprint of this manuscript has previously been published on arXiv:*******.

**Declaration of generative AI and AI-assisted technologies in the manuscript preparation process:** During the preparation of this work, the author used ChatGPT, developed by OpenAI, to assist with English-language editing, readability improvement, and manuscript organization. After using this tool, the author reviewed and edited the content as needed and takes full responsibility for the content of the published article.

## DATA AVAILABILITY

All data supporting the findings of this study are available within the main text and Appendix. One supplementary video (reference parameters) is presented.

**Video 1: Simulation time series under the baseline parameter set (Fig.2).**

https://youtu.be/OQ5RkpbYTfQ

The source code (HTML + JavaScript) used for the simulations is available on GitHub. The results presented in this paper can be reproduced by changing the UI parameters of the program.

https://github.com/Takeshi-Ishida/Self-Organizing-Swarm-Control-Model-for-Underwater-Robots

Note: In the result output files, the "Step" column records the state after each update, starting from 0; therefore, the actual number of updates corresponds to the "Step" value plus 1. In contrast, the step numbers in the image files represent the number of updates, where "step 0" corresponds to the initial configuration before any updates.

## Appendices

### Appendix 1. Overview of the Pattern Formation Model Based on Cellular Automata

### A1-1. Cellular Automaton Model for the Emergence of Turing Patterns

One of the fundamental models for reproducing biological morphogenesis is the Turing pattern, which is derived from the reaction-diffusion model proposed by Turing in 1952 [18]. This model assumes two types of substances (morphogens) with different diffusion coefficients acting as an activator and an inhibitor, and explains pattern formation through their interaction. It is expressed by the following system of partial differential equations, where $u$ and $v$ are the morphogen concentrations, $f$ and $g$ are the cross-reaction terms, and $d_1$ and $d_2$ are the diffusion coefficients:

$$\frac{\partial u}{\partial t} = d_1 \nabla^2 u + f(u,v)$$

$$\frac{\partial v}{\partial t} = d_2 \nabla^2 v + g(u,v)$$

Although Turing patterns can be obtained by numerically solving these partial differential equations, the proposed model does not use these equations directly; instead, it adopts cellular automaton models, such as those proposed by Young [23] and Ishida et al. [24], which represent the concept of local activation and long-range inhibition. In the Ishida model, which is defined on a two-dimensional cellular space, each cell $i$ takes one of two binary states (0 or 1) at time step $t$, denoted as $c_i(t) \in \{0,1\}$. The state of cell $i$ at the next time step $t+1$, denoted as $c_i(t+1)$, is determined by the sum of the states of the focal cell and its surrounding cells:

$$N_1 = \sum_{i=1}^{S_1} c_i(t) \quad , \quad N_2 = \sum_{i=1}^{S_2} c_i(t)$$

where:

$N_1$: Sum of states (0 or 1) of cells within a distance of $s_1$ grid cells from the focal cell (Figure A1-1A),

$N_2$: Sum of states (0 or 1) of cells within a distance of $s_2$ $(s_1 < s_2)$ grid cells from the focal cell (Figure A1-1A),

$S_1$: Total number of grid cells within a distance of $s_1$,

$S_2$: Total number of grid cells within a distance of $s_2$.

The state of cell $i$ at the next time step, $c_i(t+1)$, is determined by the following transition rule:

$$\text{Cell state of the next time step} = \begin{cases} 1 & : \text{ if } N_1 > N_2 \times w_1 \\ \text{Unchange} & : \text{ if } N_1 = N_2 \times w_1 \\ 0 & : \text{ if } N_1 < N_2 \times w_1 \end{cases} \qquad (A-1)$$

Here, $w_1$ is a parameter that governs the morphology of the generated pattern. Varying $w_1$ produces different Turing patterns, such as spotted or striped patterns (Figure A1-1B).

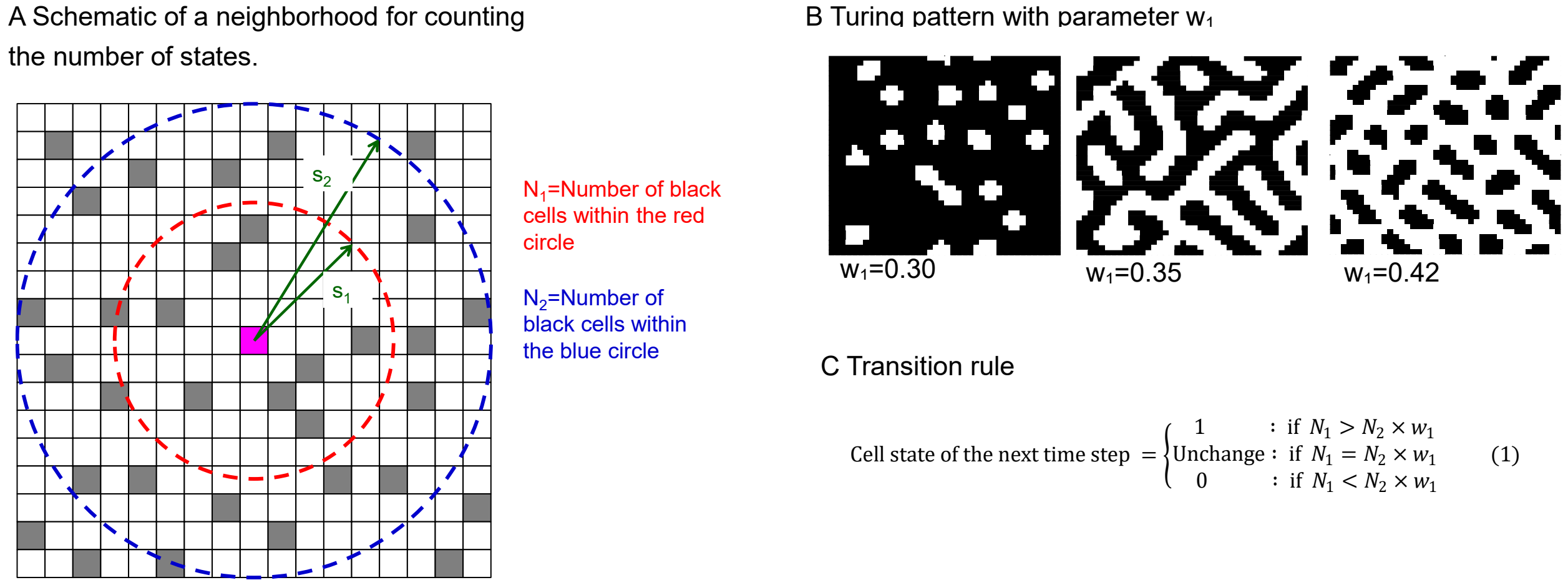

**Figure A1-1.** Formation of Turing patterns using the cellular automaton model. **(A)** Schematic diagram illustrating the summation of $N_1$ and $N_2$. Each grid cell takes a state of 0 (white) or 1 (black). The inner region represents the area within $s_1$ grid cells from the focal cell, whereas the outer region represents the area within $s_2$ grid cells. The counts of cells in state 1 within these regions correspond to $N_1$ and $N_2$, respectively. **(B)** Examples of Turing patterns generated by this model. **(C)** Transition function for computing the next cell state from $N_1$ and $N_2$.

### A1-2. The Ishida Model [19]

The patterns by the model described in A1-1 are limited to typical Turing patterns such as spots and stripes, and cannot produce complex morphologies. Ishida [21] considered a model in which, in addition to the model presented in A1-1, another parameter was added to increase the conditionals by w values. Equation (A-2) is the addition of the “0 : if $N_1 > N_2 \times w_2$” decision formula to equation (A-1). This formula adds the essence of the rule of the Conway's Game of life (survive only when there is a certain range of living cells adjacent), and only under certain conditions does the state of a cell transition to one. This model can produce a variety of forms that cannot be created by the Turing pattern. Figure A1-2 shows an example of the pattern created by this Ishida model. In this example, the calculation is initially performed from the asymmetric shape in the center, and the pattern is replicated as it moves around. Thus, a pattern emerges in which the initial asymmetry is continued and amplified. On the other hand, if calculation starts with a symmetrical pattern as the initial arrangement, a stationary pattern such as a stationary circular pattern will emerge.

The patterns generated by the model described in Section A1-1 are limited to typical Turing patterns, such as spots and stripes, and the model cannot produce complex morphologies. To overcome this limitation, Ishida [19] proposed an extended model that introduces an additional parameter to increase the number of threshold intervals governed by $w$. Equation (A-2) extends Equation (A-1) by incorporating the additional conditional rule "0: if $N_1 > N_2 \times w_2$". This modification incorporates the essence of Conway's Game of Life (in which a cell survives only when the number of adjacent living cells falls within a specific range), allowing a cell to transition to state 1 only under specific conditions. Consequently, this model can generate diverse morphologies that cannot be produced by standard Turing pattern models. Figure A1-2 shows examples of patterns generated by the Ishida model. In the first example (Figure A1-2A), the simulation starts from an asymmetric initial configuration at the center, from which the pattern splits and self-replicates while moving across the domain. Thus, a dynamic pattern emerges in which the initial asymmetry persists and is amplified over time. Conversely, when the simulation starts from a symmetric initial configuration (Figure A1-2B), stationary structures, such as a stationary circular pattern, emerge.

$$\text{Cell state of the next time step} = \begin{cases} 0 & : \text{ if } N_1 > N_2 \times w_2 \\ 1 & : \text{ if } N_1 > N_2 \times w_1 \\ \text{Unchange} & : \text{ if } N_1 = N_2 \times w_1 \\ 0 & : \text{ if } N_1 < N_2 \times w_1 \end{cases} \qquad (A-2)$$

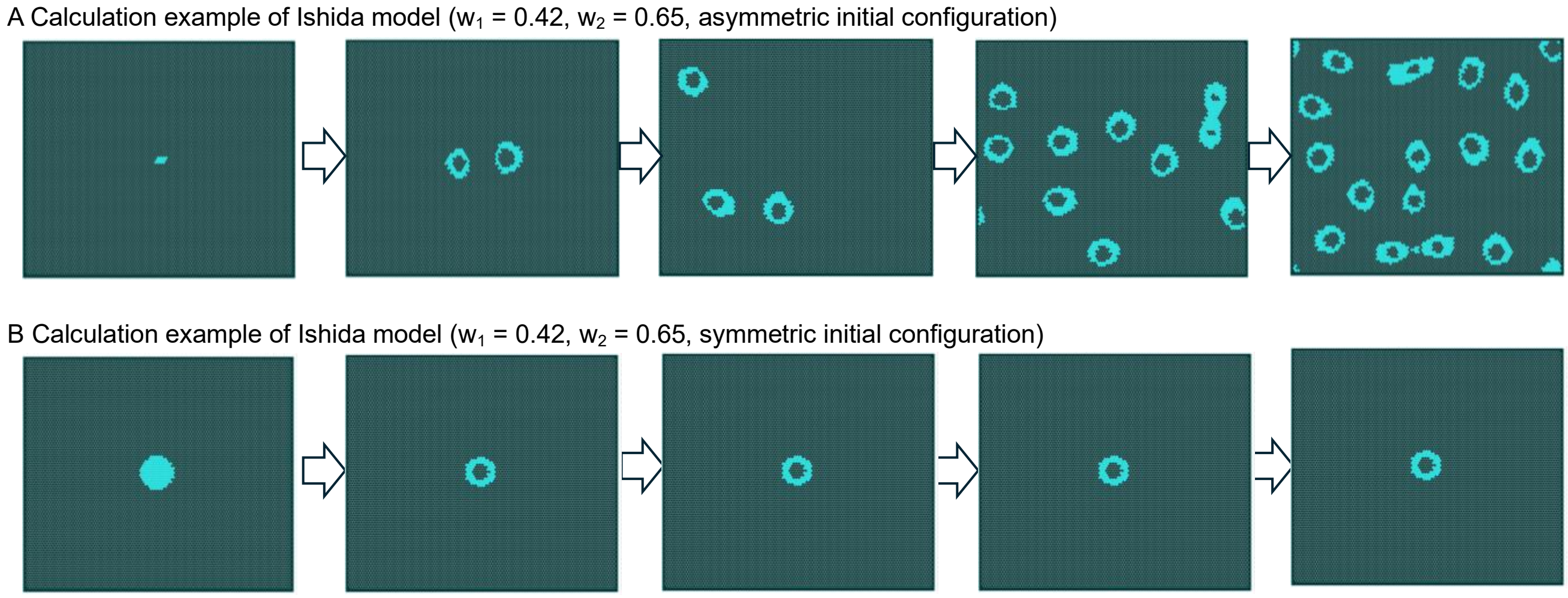

**Figure A1-2.** Simulation examples of the Ishida model. **(A)** Case producing self-replicating patterns with $w_1 = 0.42$ and $w_2 = 0.65$ from an asymmetric initial configuration. **(B)** Case producing stationary patterns with $w_1 = 0.42$ and $w_2 = 0.65$ from a symmetric initial configuration.

## Appendix 2. Algorithm for Counting the Number of Swarms

In the present study, a swarm detection algorithm based on connected component labeling was implemented to quantitatively evaluate the number of underwater robot swarms formed in simulations and the number of affiliated individuals. Although this algorithm adapts a technique used in image processing to identify spots and regions, rather than directly processing image data, it identifies swarms from the state evaluation results based on each robot's potential value and its positional coordinates in two-dimensional space. The specific computational procedure is detailed below.

**(1) Identification of Candidate Robots for Swarm Membership**

For each robot, whether it is in the stable state is evaluated using the time-averaged potential value $P_i(t)$ calculated via Eq. (3). Hysteresis is incorporated into this evaluation such that a robot in the exploratory state transitions to the stable state when $P_i(t) \geq P_{\mathrm{enter}}$, whereas a robot in the stable state reverts to the exploratory state when $P_i(t) < P_{\mathrm{exit}}$. Here, $P_{\mathrm{enter}}$ and $P_{\mathrm{exit}}$ denote the entry and exit thresholds for the stable state, respectively, satisfying $P_{\mathrm{enter}} > P_{\mathrm{exit}}$, with baseline values set to $P_{\mathrm{enter}} = 1.5$ and $P_{\mathrm{exit}} = 1.0$. Only robots classified into the stable state by this evaluation are extracted as candidate individuals for swarm membership.

**(2) Grid Mapping and Binarization of Robot Positional Data**

The computational domain ($800 \times 800$ pixels) is partitioned into square grid cells with a side length of 20 pixels, resulting in a $40 \times 40$ grid totaling 1,600 grid cells. For each robot classified into the stable state, the corresponding grid cell index is determined from its positional coordinates $(x, y)$. Each grid cell is assigned a value of 1 if at least one robot in the stable state is present within it, and 0 otherwise. This procedure represents the spatial distribution of the robots as binarized grid data. Note that even when multiple robots occupy the same grid cell, the binary cell value used for connectivity evaluation remains 1, whereas each robot is counted individually when computing the number of swarm-affiliated individuals.

**(3) Swarm Identification via Connected Component Labeling**

Connected component labeling using eight-neighbor connectivity is applied to the binarized grid data. Eight-neighbor connectivity defines adjacency across a total of eight directions surrounding each grid cell, comprising the four orthogonal directions (up, down, left, and right) and the four diagonal directions. In the algorithm, the grid cells are scanned sequentially, and whenever an unlabeled cell with a value of 1 is detected, a new swarm label is assigned to it. Subsequently, neighboring cells with a value of 1 connected to that cell are searched and assigned the same swarm label, repeating this procedure iteratively across all connected cells via a stack-based search. Through these operations, each spatially connected group of grid cells is identified as a distinct swarm. Note that periodic boundary conditions are applied to the computational domain of the proposed model; accordingly, during connectivity evaluation, the top and bottom boundaries as well as the left and right boundaries are wrapped around, such that grid cells adjacent across a boundary are classified as belonging to the same swarm.

**(4) Calculation of the Total Number of Swarms and Affiliated Individuals**

The number of connected components identified via connected component labeling is defined as the total number of swarms. Furthermore, by referencing the label of the grid cell corresponding to each robot's positional coordinates, robots in the stable state are assigned to their respective swarms, yielding the number of robots affiliated with each individual swarm. In addition, summing the affiliated individual counts across all swarms yields the total number of swarm-affiliated robots. The swarm participation rate is then calculated by dividing this total count by the total number of robots in the simulation. Note that the present implementation does not apply filtering based on a minimum area or minimum member count for connected components. Consequently, a grid cell occupied by only a single robot in the stable state can also be identified as a distinct connected component.

**(5) Calculation of Radius, Density, and Mean Potential Value for Each Swarm**

For each identified swarm, the center position is calculated from the positional coordinates of its affiliated individuals. To account for the periodic boundary conditions of the computational domain, the circular mean is employed by converting each spatial coordinate into an angular value and averaging the corresponding sine and cosine components. This approach enables accurate estimation of the center position even for swarms spanning across domain boundaries. Subsequently, the distance from each member robot to the swarm center accounting for periodic boundaries is computed, and its root mean square (RMS) is defined as the swarm radius $R$. The swarm density $D$ is calculated from the affiliated individual count $N$ and the swarm radius $R$ using the following equation:

$$D = \frac{N}{\pi \times \max(1, R)^2}$$

Here, setting a lower bound of 1 on the radius prevents division by zero when the radius equals zero. In addition, the mean potential value of each swarm is obtained by averaging the time-averaged potential values $P_i(t)$ of all robots belonging to that swarm.

## Appendix 3. Parameter Sensitivity Analysis

This appendix presents the simulation results obtained by varying each parameter individually from its baseline value, with all simulations conducted using a random seed of 12345.

### Appendix 3-1. Variation in the Number of Robots

Simulations were conducted by varying the number of robots across 40, 60, 80, 100, and 120 relative to the baseline value of 80. When the number of robots was 40, only a single swarm was formed, whereas the swarm participation rate reached its maximum when the number of robots was 100.

**Table A3-1.** Evaluation metrics under varying numbers of robots.

| Numbers of robots | Seed | w | Communication Success Rate | Command Success Rate | Number of Swarms | Number of Swarm Members | Mean Swarm Radius | Mean Swarm Density | Global Mean Potential ($\bar{P}$) | Swarm Participation Rate (%) | Command Tracking Score | Polarization (Alignment Parameter) |
|---|---|---|---|---|---|---|---|---|---|---|---|---|
| 40 | 12345 | 0.4 | 1 | 1 | 1 | 7 | 24.3 | 0.004 | 1.02 | 18 | 1.0000 | 0.999 |
| 60 | 12345 | 0.4 | 1 | 1 | 3 | 32 | 21.8 | 0.006 | 4.09 | 53 | 0.9999 | 0.999 |
| 80 | 12345 | 0.4 | 1 | 1 | 10 | 72 | 18.1 | 0.099 | 4.21 | 90 | 0.9993 | 0.994 |
| 100 | 12345 | 0.4 | 1 | 1 | 10 | 91 | 21.3 | 0.039 | 4.58 | 91 | 0.9999 | 0.999 |
| 120 | 12345 | 0.4 | 1 | 1 | 16 | 111 | 17.9 | 0.065 | 4.52 | 93 | 0.9999 | 0.999 |

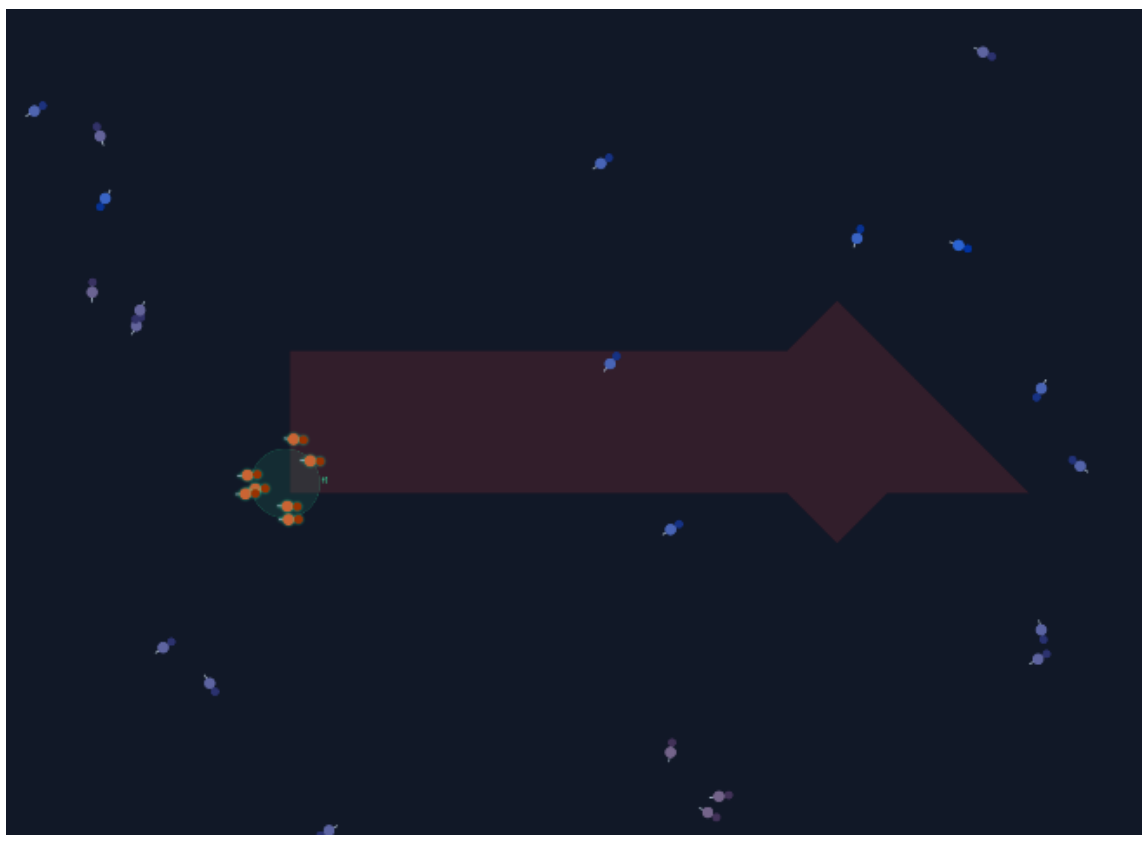

**Figure A3-1.** Simulation results at the terminal state (Step 2000) for a population of 40 robots, illustrating the difficulty of swarm formation under low robot density.

## Appendix 3-2. Variation in Mean Speed

Table A3-2 summarizes the simulation results obtained by varying the mean speed across 0.6, 0.8, 1.2, 1.4, and 1.6 relative to the baseline value of 1.2. In Table A3-2, as the mean speed increased across 0.6, 0.8, and 1.2, the overall mean potential value rose to approximately 1.8, 3.5, and 4.2, respectively. Conversely, at speeds of 1.4 and 1.6, the value was approximately 4.0 and 3.8, respectively, demonstrating that it did not increase monotonically with speed. The total number of swarms also fluctuated across these conditions, and a general optimal speed cannot be determined from the present single-seed results.

**Table A3-2.** Evaluation metrics under varying mean speeds.

| Mean speeds | Seed | w | Communication Success Rate | Command Success Rate | Number of Swarms | Number of Swarm Members | Mean Swarm Radius | Mean Swarm Density | Global Mean Potential (P̄) | Swarm Participation Rate (%) | Command Tracking Score | Polarization (Alignment Parameter) |
|---|---|---|---|---|---|---|---|---|---|---|---|---|
| 0.6 | 12345 | 0.4 | 1 | 1 | 10 | 48 | 15.6 | 0.0688 | 1.8 | 60.0 | 1.000 | 0.999 |
| 0.8 | 12345 | 0.4 | 1 | 1 | 6 | 56 | 24.4 | 0.0564 | 3.5 | 70.0 | 1.000 | 0.999 |
| 1.2 | 12345 | 0.4 | 1 | 1 | 10 | 72 | 18.1 | 0.0986 | 4.2 | 90.0 | 0.999 | 0.994 |
| 1.4 | 12345 | 0.4 | 1 | 1 | 10 | 71 | 19.4 | 0.0930 | 4.0 | 88.8 | 1.000 | 0.999 |
| 1.6 | 12345 | 0.4 | 1 | 1 | 12 | 70 | 16.6 | 0.0601 | 3.8 | 87.5 | 1.000 | 0.999 |

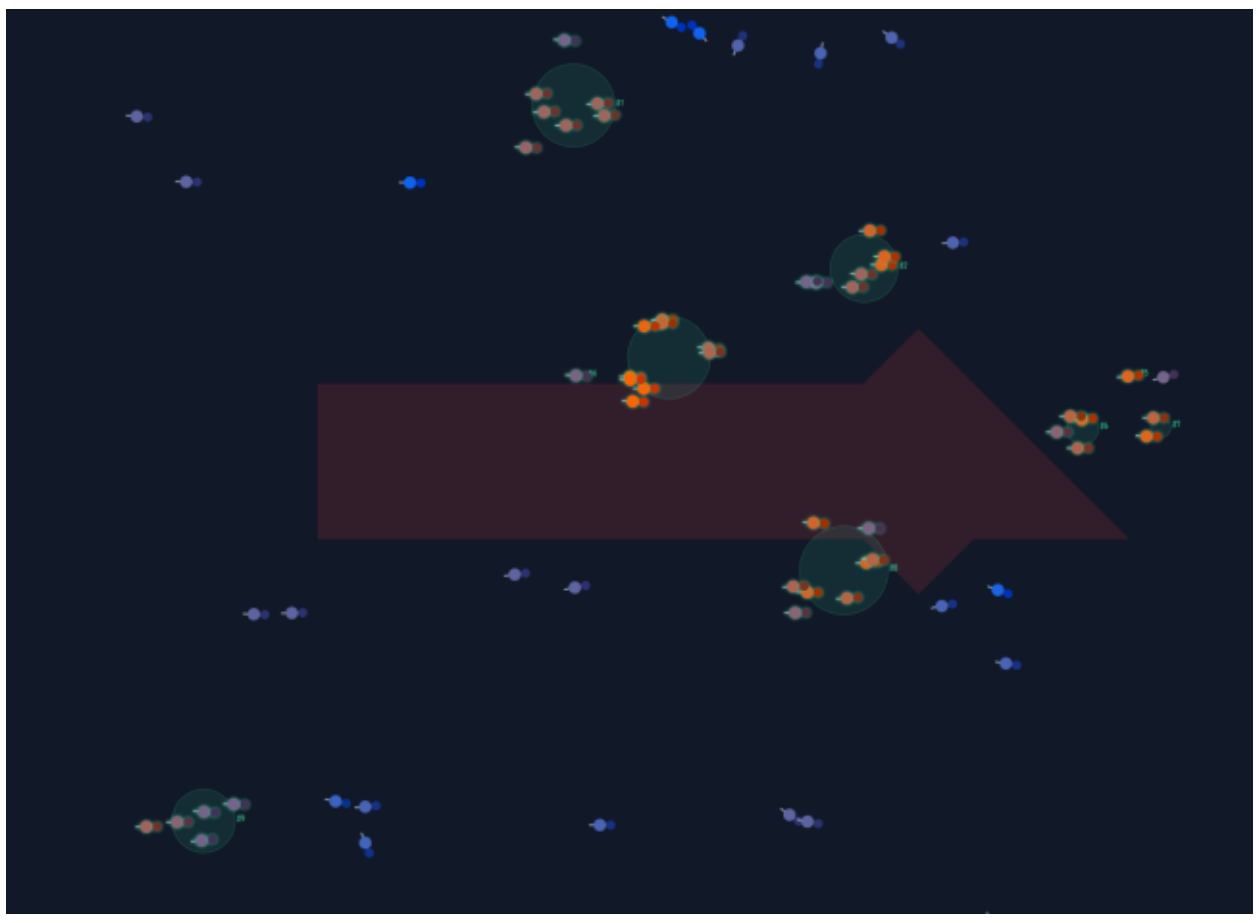

(Mean Speed 0.6)

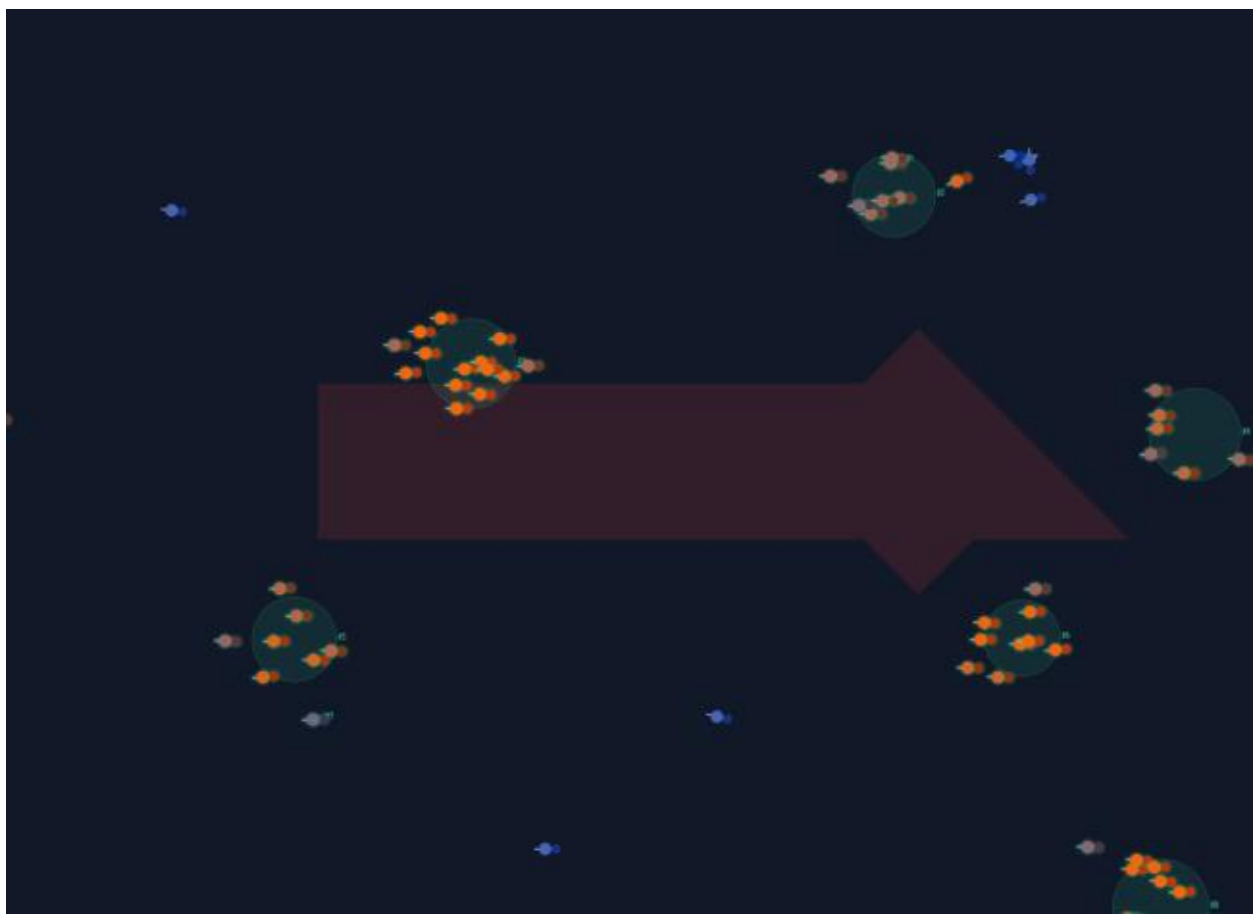

(Mean Speed 1.4)

**Figure A3-2.** Simulation results at the terminal state (Step 2000) under varying mean speeds.

## Appendix 3-3. Variation in Outer Neighborhood Range

The simulation results obtained by varying the outer neighborhood range across 80, 100, 120, 140, and 160 relative to the baseline value of 120 are presented below, with the inner neighborhood range held constant at 60.

**Table A3-3.** Evaluation metrics under varying outer neighborhood ranges.

| Sound Range (Outer) | Seed | w | Communication Success Rate | Command Success Rate | Number of Swarms | Number of Swarm Members | Mean Swarm Radius | Mean Swarm Density | Global Mean Potential (P̄) | Swarm Participation Rate (%) | Command Tracking Score | Polarization (Alignment Parameter) |
|---|---|---|---|---|---|---|---|---|---|---|---|---|
| 80 | 12345 | 0.4 | 1 | 1 | 8 | 63 | 22.8 | 0.082 | 3.66 | 78.8 | 0.9998 | 0.9992 |
| 100 | 12345 | 0.4 | 1 | 1 | 7 | 66 | 23.7 | 0.005 | 3.91 | 82.5 | 0.9997 | 0.9984 |
| 120 | 12345 | 0.4 | 1 | 1 | 10 | 72 | 18.1 | 0.099 | 4.21 | 90.0 | 0.9993 | 0.9939 |
| 140 | 12345 | 0.4 | 1 | 1 | 9 | 76 | 21.5 | 0.045 | 5.29 | 95.0 | 0.9998 | 0.9991 |
| 160 | 12345 | 0.4 | 1 | 1 | 7 | 73 | 21.5 | 0.050 | 5.81 | 91.3 | 0.9998 | 0.9991 |

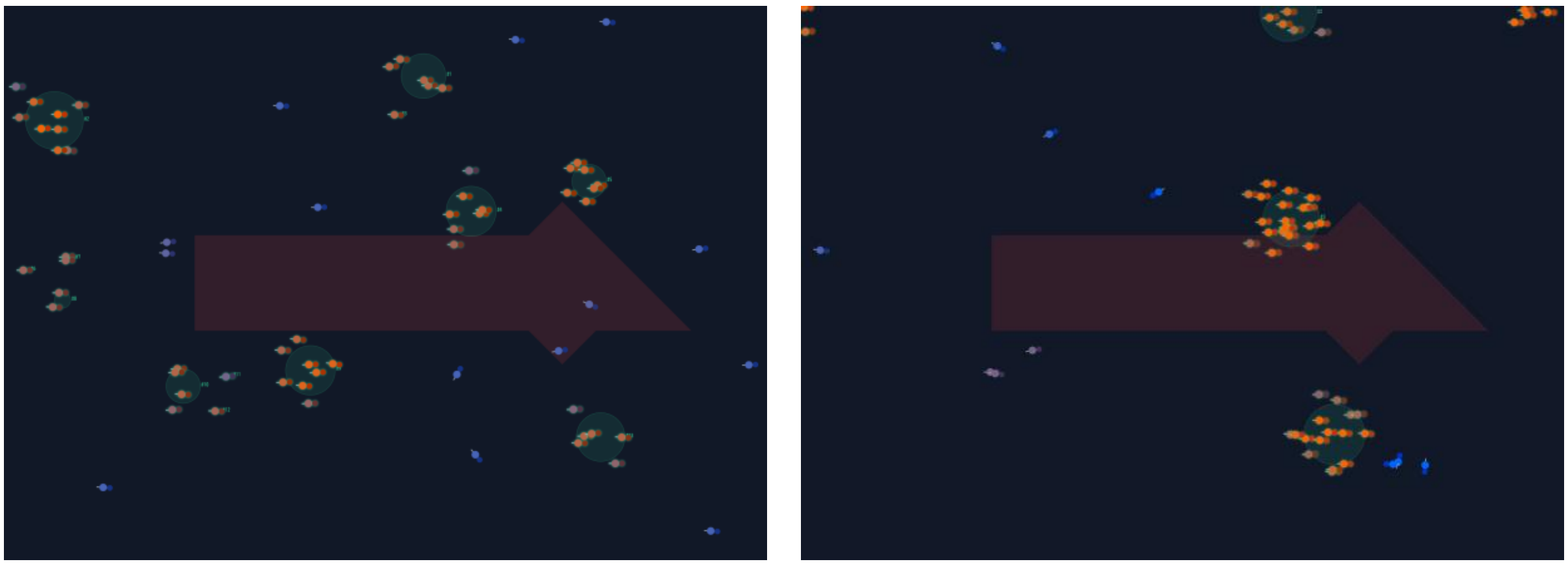

(Outer 80) (Outer 160)

**Figure A3-3.** Simulation results at the terminal state (Step 2000) under varying outer neighborhood ranges.

## Appendix 3-4. Variation in Inner Neighborhood Range

The simulation results obtained by varying the inner neighborhood range across 20, 40, 60, 80, and 100 relative to the baseline value of 60 are presented below, with the outer neighborhood range held constant at 120.

**Table A3-4.** Evaluation metrics under varying inner neighborhood ranges.

| Radio Range (Inner) | Seed | w | Communication Success Rate | Command Success Rate | Number of Swarms | Number of Swarm Members | Mean Swarm Radius | Mean Swarm Density | Global Mean Potential (P̄) | Swarm Participation Rate (%) | Command Tracking Score | Polarization (Alignment Parameter) |
|---|---|---|---|---|---|---|---|---|---|---|---|---|
| 20 | 12345 | 0.4 | 1 | 1 | 4 | 12 | 4.6 | 0.169 | 0.1426 | 15.0 | 0.8598 | 0.9941 |
| 40 | 12345 | 0.4 | 1 | 1 | 7 | 59 | 15.2 | 0.054 | 3.6752 | 73.8 | 0.9896 | 0.9993 |
| 60 | 12345 | 0.4 | 1 | 1 | 10 | 72 | 18.1 | 0.099 | 4.2142 | 90.0 | 0.9993 | 0.9939 |
| 80 | 12345 | 0.4 | 1 | 1 | 29 | 75 | 8.7 | 0.163 | 3.3651 | 93.8 | 0.9995 | 0.9997 |
| 100 | 12345 | 0.4 | 1 | 1 | 56 | 80 | 3.7 | 0.237 | 3.2286 | 100.0 | 0.9992 | 0.9999 |

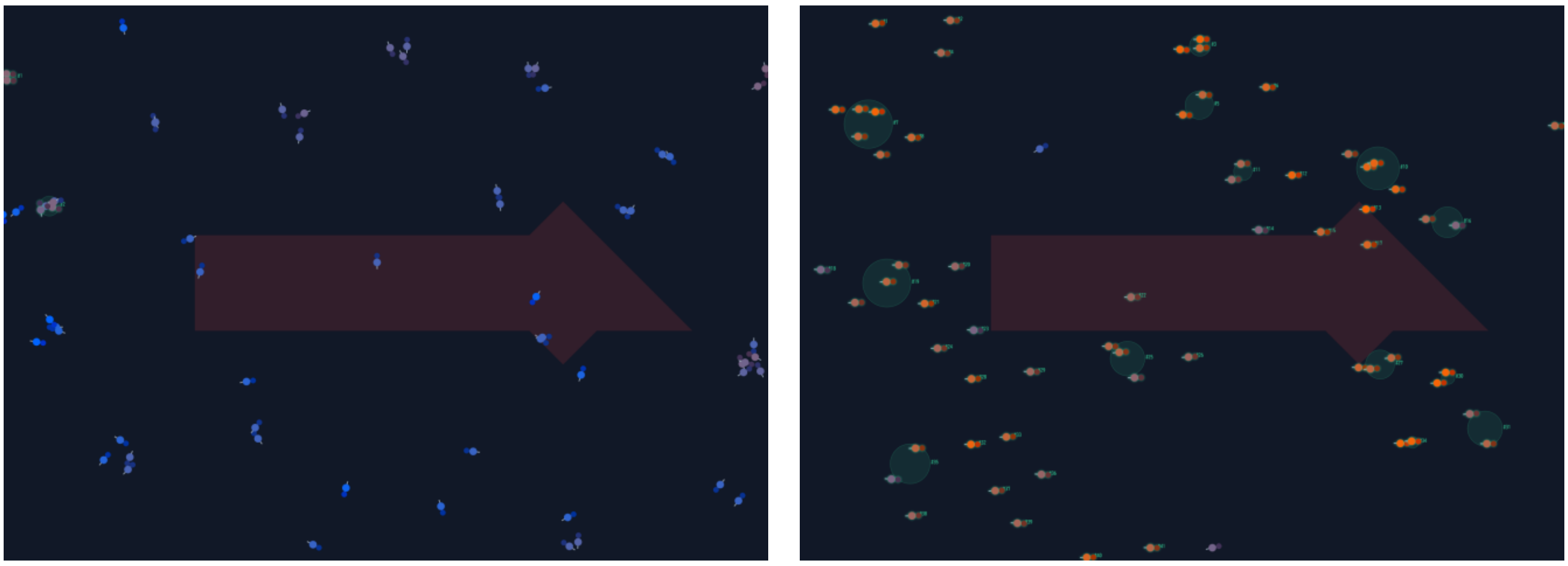

(Inner 20) (Inner 100)

**Figure A3-4.** Simulation results at the terminal state (Step 2000) under varying inner neighborhood ranges.

**Appendix 3-5. Variation in Command Direction**

The simulation results obtained by varying the command direction across 0°, 30°, 60°, and 90° relative to the baseline value of 0° are presented below. Differences in the evaluation metrics were observed depending on the command direction, which is attributed to stochastic variations arising from factors such as initial vehicle placement. Therefore, future work must conduct simulations across multiple random seeds to decouple and evaluate directional dependence and stochastic variations separately.

**Table A3-5.** Evaluation metrics under varying command directions.

| Command Angle | Seed | w | Communication Success Rate | Command Success Rate | Number of Swarms | Number of Swarm Members | Mean Swarm Radius | Mean Swarm Density | Global Mean Potential ($\bar{P}$) | Swarm Participation Rate (%) | Command Tracking Score | Polarization (Alignment Parameter) |
|---|---|---|---|---|---|---|---|---|---|---|---|---|
| 0 | 12345 | 0.4 | 1 | 1 | 10 | 72 | 18.1 | 0.099 | 4.21 | 90.0 | 0.9993 | 0.9939 |
| 30 | 12345 | 0.4 | 1 | 1 | 5 | 70 | 28.1 | 0.005 | 5.83 | 87.5 | 0.9998 | 0.9991 |
| 60 | 12345 | 0.4 | 1 | 1 | 6 | 68 | 27.1 | 0.005 | 4.73 | 85.0 | 1.0000 | 0.9986 |
| 90 | 12345 | 0.4 | 1 | 1 | 11 | 78 | 20.3 | 0.042 | 4.36 | 97.5 | 0.9997 | 0.9991 |

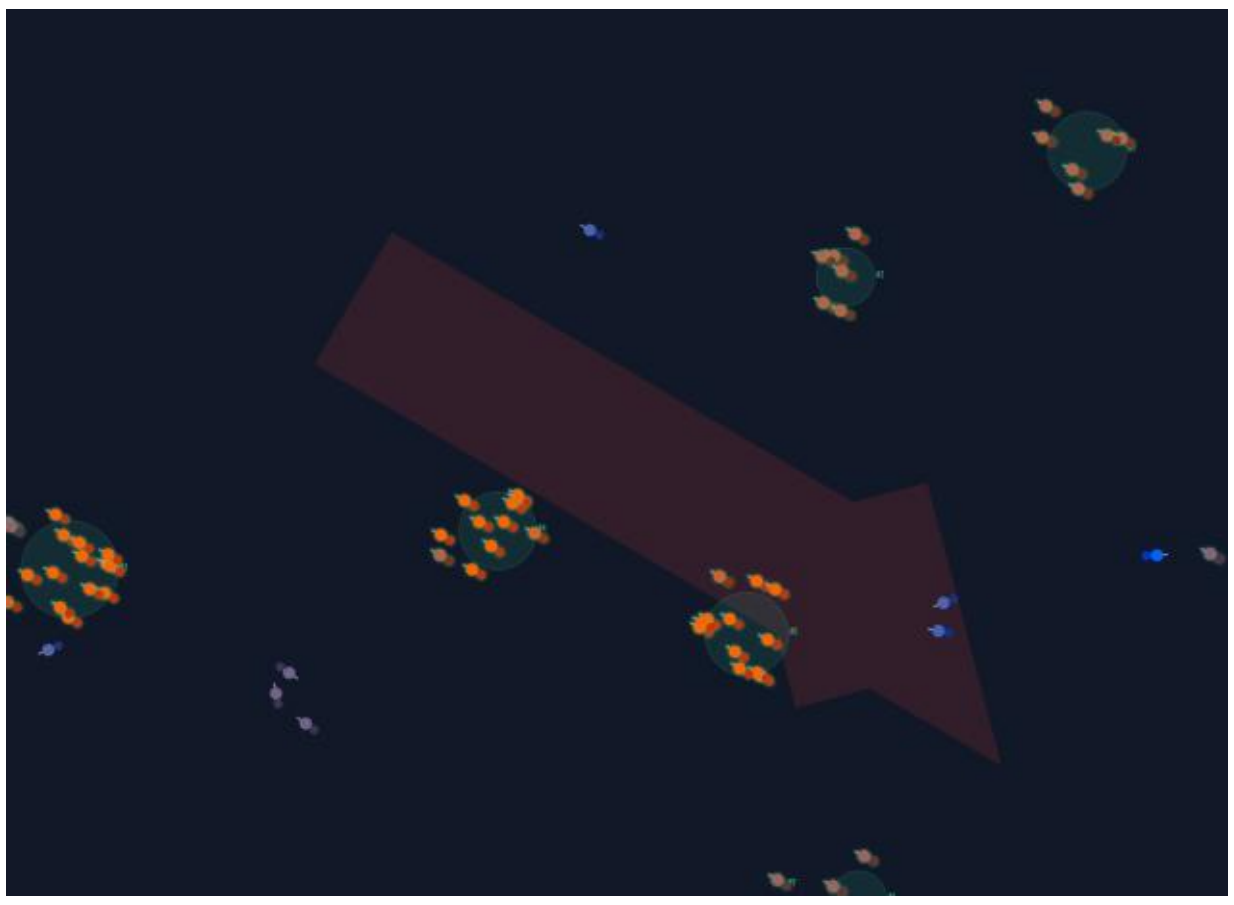

(Command Angle 30°)

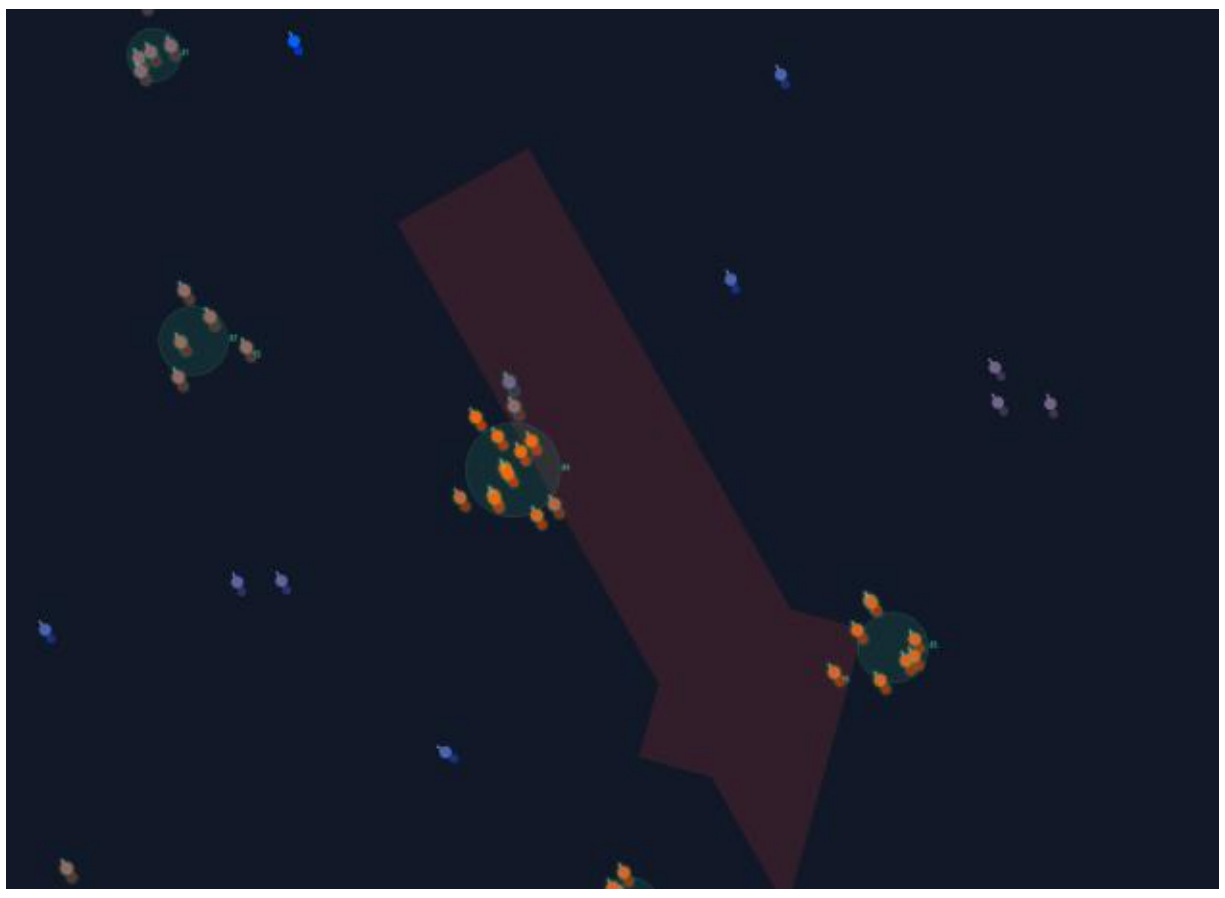

(Command Angle 60°)

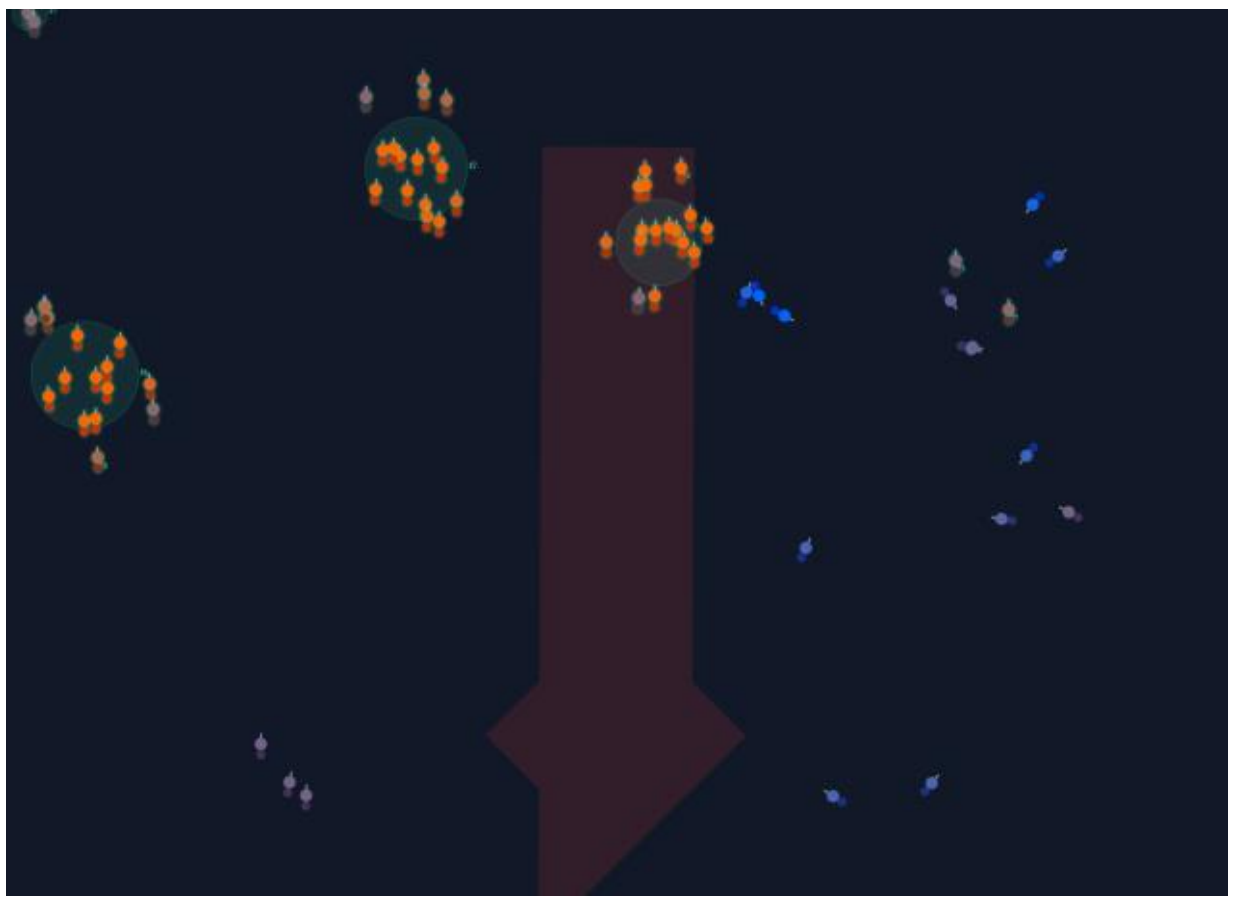

(Command Angle 90°)

**Figure A3-5.** Simulation results at the terminal state (Step 2000) under varying command directions.

## Appendix 3-6. Variation in Mixing Weight

The simulation results obtained by varying the mixing weight $W_{mix}$ across 0.1, 0.2, 0.3, and 0.4 relative to the baseline value of 0.2 are presented below.

**Table A3-6.** Evaluation metrics under varying mixing weights.

| Mix Weight | Seed | w | Communication Success Rate | Command Success Rate | Number of Swarms | Number of Swarm Members | Mean Swarm Radius | Mean Swarm Density | Global Mean Potential (P̄) | Swarm Participation Rate (%) | Command Tracking Score | Polarization (Alignment Parameter) |
|---|---|---|---|---|---|---|---|---|---|---|---|---|
| 0.1 | 12345 | 0.4 | 1 | 1 | 15 | 80 | 15.4 | 0.055 | 3.98 | 100.0 | 0.9999 | 0.9995 |
| 0.2 | 12345 | 0.4 | 1 | 1 | 10 | 72 | 18.1 | 0.099 | 4.21 | 90.0 | 0.9993 | 0.9939 |
| 0.3 | 12345 | 0.4 | 1 | 1 | 8 | 59 | 17.2 | 0.046 | 4.26 | 73.8 | 0.9998 | 0.9995 |
| 0.4 | 12345 | 0.4 | 1 | 1 | 10 | 65 | 19.7 | 0.036 | 3.57 | 81.3 | 0.9998 | 0.9992 |

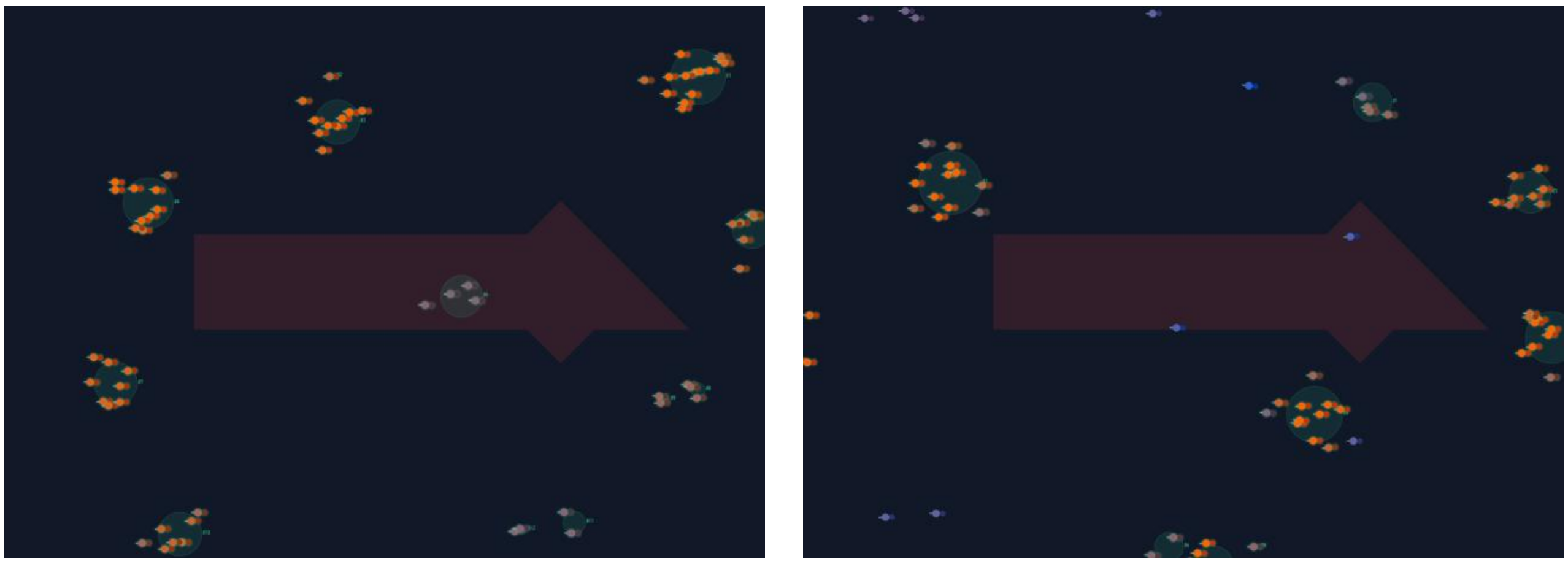

（Mix Weight 0.1）　　（Mix Weight 0.4）

**Figure A3-6.** Simulation results at the terminal state (Step 2000) under varying mixing weights.

## Appendix 3-7. Variation in Time-Averaging Coefficient $\alpha$

The simulation results obtained by varying the potential time-averaging coefficient $\alpha$ $(0 < \alpha \leq 1)$ from 0.1 to 0.4 relative to the baseline value of 0.2 are presented below.

**Table A3-7.** Evaluation metrics under varying time-averaging coefficients $\alpha$.

| α | Seed | w | Communication Success Rate | Command Success Rate | Number of Swarms | Number of Swarm Members | Mean Swarm Radius | Mean Swarm Density | Global Mean Potential (P̄) | Swarm Participation Rate (%) | Command Tracking Score | Polarization (Alignment Parameter) |
|---|---|---|---|---|---|---|---|---|---|---|---|---|
| 0.1 | 12345 | 0.4 | 1 | 1 | 8 | 69 | 19.2 | 0.123 | 5.22 | 86.3 | 1.000 | 0.999 |
| 0.2 | 12345 | 0.4 | 1 | 1 | 10 | 72 | 18.1 | 0.099 | 4.21 | 90.0 | 0.999 | 0.994 |
| 0.3 | 12345 | 0.4 | 1 | 1 | 10 | 70 | 20.2 | 0.037 | 3.95 | 87.5 | 1.000 | 0.999 |
| 0.4 | 12345 | 0.4 | 1 | 1 | 11 | 70 | 15.4 | 0.119 | 4.31 | 87.5 | 1.000 | 0.999 |